\documentclass{svproc}
\usepackage{url}

\usepackage[sort&compress,square,numbers]{natbib}
\usepackage{graphicx}
\usepackage{graphbox}
\usepackage{multicol}
\usepackage[bottom]{footmisc}
\usepackage{siunitx}
\usepackage{tabulary}
\usepackage{amsmath}
\usepackage{mathtools}
\usepackage[export]{adjustbox}

\usepackage{algorithm}
\usepackage{algpseudocode}

\usepackage{hyperref}
\hypersetup{
    colorlinks=true,
    linkcolor=blue,
    citecolor=blue,
    filecolor=magenta,      
    urlcolor=cyan,
    pdftitle={Overleaf Example},
    pdfpagemode=FullScreen,
    }

\let\vec\mathbf

\usepackage[subtle,trackingfraction=0.995]{savetrees}
\expandafter\def\expandafter\normalsize\expandafter{%
    \normalsize%
    \setlength\abovedisplayskip{4pt}%
    \setlength\belowdisplayskip{4pt}%
    \setlength\abovedisplayshortskip{3pt}%
    \setlength\belowdisplayshortskip{6pt}%
}

\begin{document}
\mainmatter              
\title{Distributed Spatial Awareness for Robot Swarms}
\author{Simon Jones \and Sabine Hauert}

\institute{University of Bristol, Bristol, UK\\
\email{simon2.jones@bristol.ac.uk}, \email{sabine.hauert@bristol.ac.uk}
}

\maketitle              

\begin{abstract}
Building a distributed spatial awareness within a swarm of locally sensing and communicating robots enables new swarm algorithms. We use local observations by robots of each other and Gaussian Belief Propagation message passing combined with continuous swarm movement to build a global and distributed swarm-centric frame of reference. With low bandwidth and computation requirements, this shared reference frame allows new swarm algorithms. We characterise the system in simulation and demonstrate two example algorithms.

\end{abstract}
\section{Introduction}
Swarm robotics, inspired by swarms in nature, has the potential for resilient, robust, and redundant solutions to a wide range of problems such as mapping, logistics, search and rescue, disaster recovery, and environmental monitoring. Many relatively simple and cheap robots, each following simple rules, with local interactions between themselves and the environment are capable of producing a desired emergent swarm-level behaviour \cite{schranz2020swarm,jones2016evolving,crosscombe2017robust,birattari2019automatic,tzoumas2023wildfire,jones2019onboard}. 

There are many areas of swarm algorithm design where access to global information would be useful, but unless that information is inferred or constructed through purely local interactions, it does not fit within the distributed swarm paradigm. One example is spatial awareness, by which we mean that an agent within the swarm is aware of its own location with respect to the swarm as a whole; the swarm shares a spatial reference frame. The availability of a completely distributed, completely local, low cost, shared reference frame would open up algorithmic approaches previously not possible. By using Gaussian Belief Propagation, we can construct this within a swarm of robots based only on local observation and messaging. Robots move around, constructing a distributed, size-limited factor graph of observations of other robots and odometry information. Message passing within and between neighbouring robots results in convergence on a shared frame of reference; each robot knows where it is in relation to it. To demonstrate the potential of this, we focus on two applications, shape formation and logistics

Swarm shape formation is a proxy for a number of real-world problems such as search and rescue or emergency communication. Many problems rely on the swarm maintaining a coherent shape or coverage of particular areas. This has been tackled in swarm robotics in a number of ways which in their essence involve the construction of a frame of reference, or coordinate system. These systems often rely on robots transitioning to a static state to serve as anchors for further extensions to the shape and coordinate system, or unrealistic assumptions of position knowledge. 

The use of swarms for intralogistics is an emerging area, where perhaps we can move beyond the lab into real-world applications. With our DOTS \cite{jones2022dots} robots we aim to demonstrate a simple but functional application. 
Specifically, we consider the potential for small scale out-of-the-box solutions for everyday environments; 
 simply delimit an area of floor and add robots and small cargo carriers. Users would download an app to their phone and use this to call for a carrier to deposit an item. Via Bluetooth, any robot within range would respond and provide the carrier and take it and the item to be stored. To retrieve the item, the user would use the app again, robots would talk with their neighbours until a robot with recent knowledge of the item heard, which would pick up the carrier and take it to the user. Even simple random walk algorithms are capable of effective retrieval in logistics applications.

We describe the implementation of a system, which we call Distributed Spatial Awareness (DSA), to provide a completely local and distributed shared frame of reference. We characterise its performance in simulation, examine the trade-offs with computational and communication cost, and to demonstrate it, we show simple but effective shape formation, and enhanced knowledge awareness within an intralogistics application. This paper is organised as follows, in the next section, we look at the background, Section \ref{sec:methods} covers methods, Section \ref{sec:results} analyses and discusses the results, and Section \ref{sec:conc} concludes.
\section{Background}

\subsection{Gaussian Belief Propagation}
Gaussian Belief Propagation (GBP) is a method of performing distributed iterative probabilistic inference or state estimation on a graph of relations between Gaussian variables by means of message passing. It is not new \cite{pearl1982reverend,weiss1999correctness,bickson2008distributed} but has received recent attention, with convergence guarantees \cite{su2015convergence,du2017convergence} and applicability to distributed systems \cite{davison2019futuremapping, ortiz2020bundle,murai2022robot, patwardhan2022distributing}. A widely used way to represent this inference problem is a \textit{factor graph} \cite{dellaert2017factor}; a bipartite graph consisting of nodes which are either variables or factors. Variables are quantities we wish to estimate or infer and factors, which connect to one or more variables, represent constraints on those connected variables, for example derived from some measurement of the environment.

Performing inference on a factor graph is well studied and at the heart of many robotics estimation problems. Solver libraries such as Ceres \cite{agarwal2022ceres} and GTSAM \cite{dellaert2022gtsam} are highly performant, but rely on centralised representation and processing. \citet{davison2019futuremapping} make the argument that the completely decentralised and incremental processing of GBP is a better fit for future systems as processing power becomes more distributed. This naturally fits with the swarm robotics paradigm, and allows the inference of global properties by using purely local interactions.

In this work, we use only linear 2D state representations, and only factors of one or two variables. Gaussian variables $\vec{x_i}$ in the moments form $\mathcal{N}(\vec{x_i};\mu_i,\Sigma_i)$ can also be represented in the canonical/information form $\mathcal{N}^{-1}(\vec{x_i};\eta_i,\Lambda_i)$, connected by the identities: $\Lambda=\Sigma^{-1},\eta=\Lambda\mu$, where $\Lambda$ is the precision matrix and $\eta$ the information vector. Hereafter, when specifying components, we use the notation $(\eta,\Lambda)_G\equiv(\mu,\Sigma)_G$. Factor $f_j(\vec{x_j})$ connects to a single variable and specifies a prior, defined by the Gaussian constraint $\vec{z_j}=(\eta_j,\Lambda_j)_G$. Factor $f_k(\vec{x_{k1}},\vec{x_{k2}})$ connects two variables $\vec{x_{k1}}, \vec{x_{k2}}$ and specifies a measurement, comprising the functional relationship $\vec{h_k}(\vec{x_{k1}}, \vec{x_{k2}})=\vec{x_{k2}}-\vec{x_{k1}}$, and the constraint $\vec{z_k}=(\eta_k,\Lambda_k)_G$. As described in more detail in \cite{davison2019futuremapping,ortiz2021visualgbp}, the GBP algorithm requires three steps: \textbf{factor-to-variable message} $m_{f\to x}$, \textbf{variable-to-factor-message} $m_{x\to f}$, and \textbf{belief update} $b(x)$. All messages are in the form $(\eta,\Lambda)_G$. The schedule of operations, and nodes upon which the operations take place, can be entirely arbitrary and asynchronous, though as noted in \cite{ortiz2021visualgbp} convergence time is affected by the form of the schedule. 

\textbf{Belief update:} A variable $\vec{x_i}$ has its belief updated as the product of messages from all connected factors. In canonical form, this is expressed as a sum: 
\begin{align}
b(\vec{x_i})=\left(\sum_{f\in n(\vec{x_i})}{\eta}_{f\to x},\sum_{f\in n(\vec{x_i})}{\Lambda_{f\to x}}\right)_G\label{eqn:bel}
\end{align} 
where $n(\vec{x_i})$ are all the factors connected to $\vec{x_i}$.

\textbf{Variable-to-factor message:} The message to a connected factor is the product of the incoming messages from all other connected factors:
\begin{align}
	m_{x\to f_j}=\left(\sum_{f\in n(\vec{x_i})\setminus f_j}{\eta}_{f\to x},\sum_{f\in n(\vec{x_i})\setminus f_j}{\Lambda_{f\to x}}\right)_G
\end{align}

\textbf{Factor-to-variable message:} For a single variable factor, this is simply the factor. 
\begin{align}
	f(\vec{x}):m_{f\to x} &= \vec{z} 
\end{align}
For a two variable measurement factor, this is the product of the factor and the message from the other variable, marginalising out the other variable. Alternatively, the message to a variable is the precision weighted sum of the message from the other variable and the measurement vector $\vec{z}$.
\begin{align}
\text{Let }\alpha_{p} &= \frac{\Lambda_f}{\Lambda_f+\Lambda_p} \label{eqn:alpha} \\
f(\vec{x_i},\vec{x_j}):m_{f\to x_i} &= \left((1-\alpha_{x_j})\eta_f+\alpha_{x_j}\eta_{x_j}),\alpha_{x_j}\Lambda_{x_j}\right)
&\text{message to } \vec{x_i} \label{eqn:m1} \\
f(\vec{x_i},\vec{x_j}):m_{f\to x_j} &= \left((1-\alpha_{x_i})\eta_f+\alpha_{x_i}\eta_{x_i}),\alpha_{x_i}\Lambda_{x_i}\right)
&\text{message to } \vec{x_j} \label{eqn:m2}
\end{align}
One modification we make to the original algorithm is to note that: 
\begin{align}
m_{x\to f_j} = b(\vec{x_i}) - m_{f_j\to x}	
\end{align}
Variable nodes just calculate their beliefs and send them as messages, and factor nodes locally calculate what the variable-to-factor message would have been by subtracting the last message sent to that variable. This minimises the non-local knowledge needed in any node.

\textbf{Message damping} As noted in \cite{su2015convergence}, message damping often improves convergence. We only damp messages from factors to variables, such that the message components $(\eta,\Lambda)$ are replaced:
\begin{align}
	\eta'_{t+1} &= (1-r_{damp})\eta_{t+1} + r_{damp}\eta_{t} \\
	\Lambda'_{t+1} &= (1-r_{damp})\Lambda_{t+1} + r_{damp}\Lambda_{t}
\end{align}
Throughout this work, we use a damping factor of $r_{damp}=0.8$, arrived at empirically during initial experimentation as a compromise between slow convergence and a tendency to oscillation.

\subsection{Shape formation}
Shape or pattern formation in robot swarms is widely studied. The use of a swarm to perform a task is often implicitly or explicitly dependent upon that swarm maintaining a particular shape or covering a particular area. Often, a necessary part of forming a shape is the use of simpler behaviours such as dispersion, aggregation, alignment. Notably \citet{reynolds1987flocks} introduced simple rules to produce complex flocking behaviour.
Potential fields or virtual springs are a common approach \cite{pan2019virtual,sabattini2011arbitrarily} but often use assumptions such as knowledge of position.  \citet{rubenstein2014programmable} used kilobots \cite{rubenstein2012kilobot} to build arbitrary 1-connected shapes by individual robots extending a seed formation and localising themselves against already positioned static robots, likewise \citet{li2019decentralized} progressively construct a shape with new agents positioning themselves against those already present.
Also there are bioinspired approaches such as morphogenesis \cite{carrillo2019toward,slavkov2018morphogenesis}. 
See \citet{oh2017bio} for a recent review.

\subsection{Swarm logistics}
Swarm logistics can be regarded as a real-world application of foraging \cite{winfield2009foraging,liu2010modeling,pitonakova2018recruitment,talamali2020sophisticated}. Robots must leave a \textit{nest} area and seek resources to be returned to the nest. The analogy is obvious. As a use case, we focus on robot swarms used for intralogistics. As noted above, we specifically focus on the out-of-the-box, easily deployable solutions for everyday environments, as described in \cite{jones2020distributed}. Even simple random walk algorithms are capable of effective retrieval in logistics applications, \cite{milner2022stochastic,milner2022swarm}. 
Messages from users propagate from robot to robot, robots store and retrieve items, all in parallel and without central resources. Although simple, this scenario encapsulates many issues that will need to be solved for viable swarm logistics systems to become a reality.

\section{Methods}\label{sec:methods}
We work in simulation, with the intention of transferring the work to real robots in the future. As such, we use an abstract model of our real robots, the DOTS \cite{jones2022dots}; each robot is modelled as a \SI{2}{kg} disk, \SI{250}{mm} in diameter, that can move holonomically at up to $v_{max}=$\,\SI{1}{ms^{-1}}, and can sense and identify other robots and objects to a distance of $r_{sense}=$\,\SI{0.5}{m}. Communication is possible between any robots that can sense each other. Each robot has imperfect sensors distorted by Gaussian noise; a velocity sensor $\vec{v_{sense}}=GT(v_{robot}) \cdot \mathcal{N}(1,\sigma^2_{velocity})$, and a relative position sensor $\vec{p_{object}}=GT(object)-GT(robot)+\mathcal{N}(0,\sigma^2_{position})$ where $GT(k)$ is the ground truth from the simulator. The robot maintains odometry $\vec{p_{odom}} = \sum \vec{v_{sense}}$, integrated since the last variable node was created. 

\subsection{Simulator}
The simulator is written in C++, using the Box2D \cite{catto2011box2d} physics engine. Updates to the simulation occur at \SI{60}{Hz}. The arena is a square area with fixed walls. Robot motion is modelled as a disk with friction sliding on the arena surface, collisions between robots and with walls follow physics. Low level proportional control of the force applied to the robot body is used to satisfy the commanded velocity. As well as robots, there can be objects that can be detected by the robots. For each robot a list of robots and objects within $r_{sense}$ range is maintained, and at each timestep a set of abstract senses is constructed and a controller routine is executed on those senses to generate a commanded velocity. The simulated senses and actuator are shown in Table \ref{tab:senses}.
\begin{table}[h]\centering\caption{Robot senses and actuator}\label{tab:senses}
\begin{tabulary}{\columnwidth}{ll}
	\hline
	Senses & Description \\
	\hline
	$\vec{v_{sense}}$ & Velocity  \\
	$\vec{p_{odom}}$ & Odometry since last variable node \\
	$\vec{p_{robot}}=\vec{x_i}+\vec{p_{odom}}$ & Inferred position of robot relative to swarm centroid\\
	$\{\vec{p^1_{object}},..,\vec{p^n_{object}}\}$ & List of other robots and objects within $r_{sense}$ radius \\
	$\angle_{neighbours}$ & Direction of nearest neighbour robots \\
	$converged$ & True when elapsed time $>t_{proxyconv}$\\
	\hline
	Actuator &  \\
	\hline
	$\vec{v_{cmd\_vel}}$ & Commanded velocity of robot\\
	\hline
\end{tabulary}
\end{table}

The baseline controller performs a random walk moving at $v_{fast}=0.5$\SI{}{ms^{-1}} in a random direction $t_{rwduration}=\max(0.1,\mathcal{N}(2,1))$\,seconds, before choosing a new direction and duration. Collisions are treated as ballistic. This behaviour is denoted DSA-RW (Distributed Spatial Awareness - Random Walk).

\begin{figure}
	\centering
	\includegraphics[width=0.5\columnwidth]{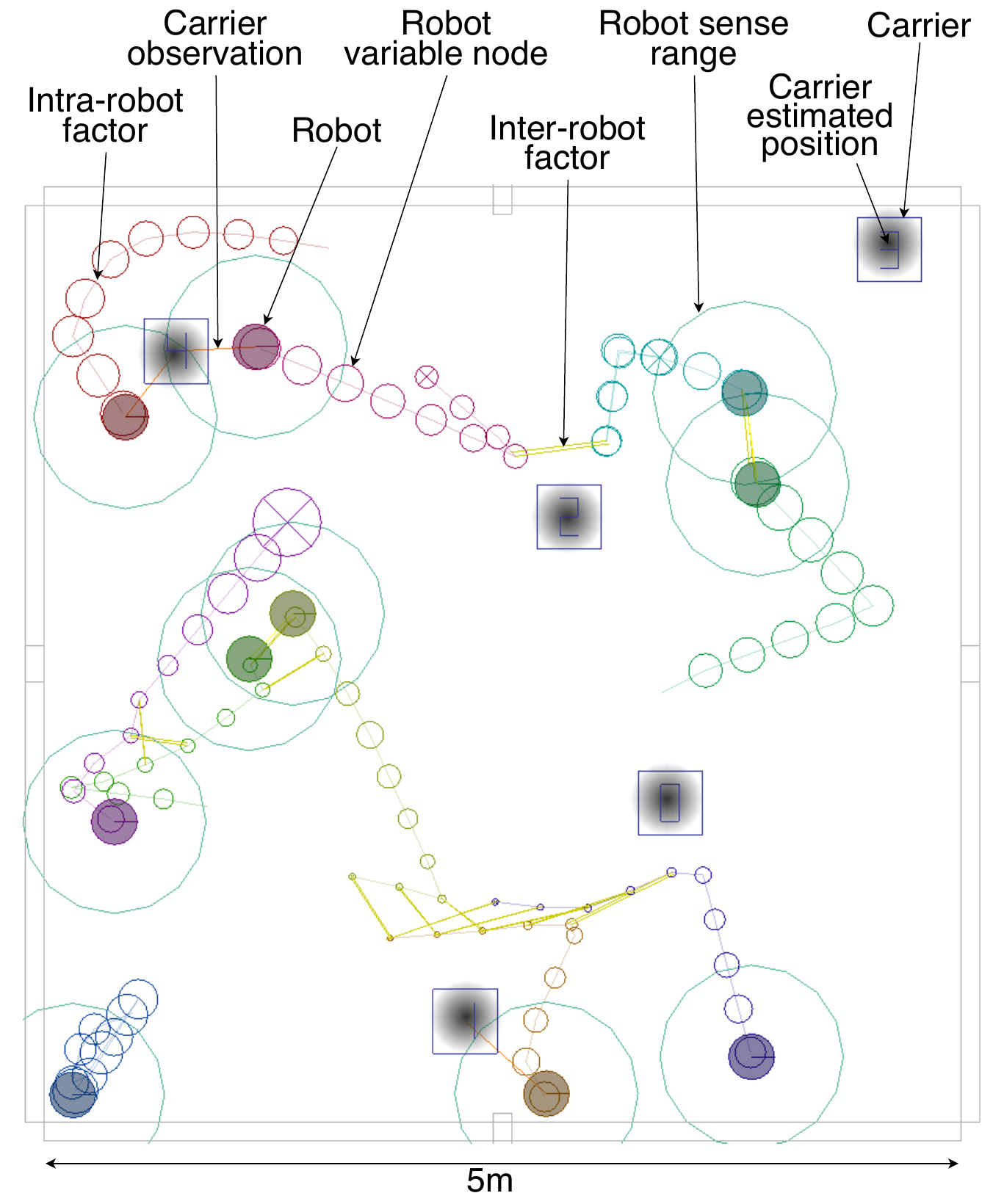}
	\caption{Simulation of robots performing DSA-RW to locate cargo carriers, with various elements of the visualisation labelled. Within the carrier squares are overlayed the current swarm estimates of the carrier position, already showing good correspondence after two minutes of simulated time.}\label{fig:sim}
\end{figure}

\subsection{Distributed Spatial Awareness}
We construct a shared reference frame for the swarm in the following way: Each robot builds a local factor graph with variables representing 2D pose, unary anchor factors connecting to a single variable, and binary relative measurement factors connecting two variables. 
Each robot shares synchronised time, and at regular intervals $t_{node}$ from a starting epoch a new timestep $ts_i$ is issued and a new variable node $\vec{x_i}$ is created. At any given time, the pose of the robot relative to the shared reference frame is $\vec{p_{robot}} = \vec{x_i} + \vec{p_{odom}}$. 

The first node to be created will have an anchor factor connected to it with a weak prior of pose $(0,0)$. Successive nodes have a measurement factor connecting them:   $f_{\vec{x_i}\to \vec{x_{i-1}}}(\mu=\vec{p_{odom}},\Sigma=\sigma^2_{velocity}I_2)$. As new variable nodes are created, old ones are removed to maintain a maximum number of nodes $n_{window}$, with the now oldest variable having an anchor factor attached to it set to the belief of that variable. At every $t_{message}$ interval, a factor node is chosen at random, and messages are propagated to each connected variable according to the GBP algorithm. Each connected variable then has its belief updated.

As described, this local factor graph is purely a bounded time window sampling of the odometry of the robot. In order to connect the local factor graphs together into a swarm whole, we add two things: 1) Robots observe other robots within their sensory range and create measurement factors $f(\mu=\vec{p_{object}},\Sigma=\sigma^2_{position}I_2 )$ that link the current variable node on the observing robot to the observed robot. These factors, termed outward-facing, are given the current timestep $ts_i$ and the ID of the other robot. This uniquely identifies the variable at the other robot that it links to. 2) Robots in sensory range exchange GBP messages. At the same $t_{message}$ interval as above, the robot chooses a single other robot from any within $r_{sense}$ range and sends a message request. This consists of a list of the timesteps of all outward-facing factors that connect to the other robot. The responding robot replies with a list of beliefs from the variables those factors uniquely connect to, and a list of messages from remote factors that connect to local variables of the requesting robot.

\begin{figure}
	\centering
	\includegraphics[align=c,width=0.49\linewidth]{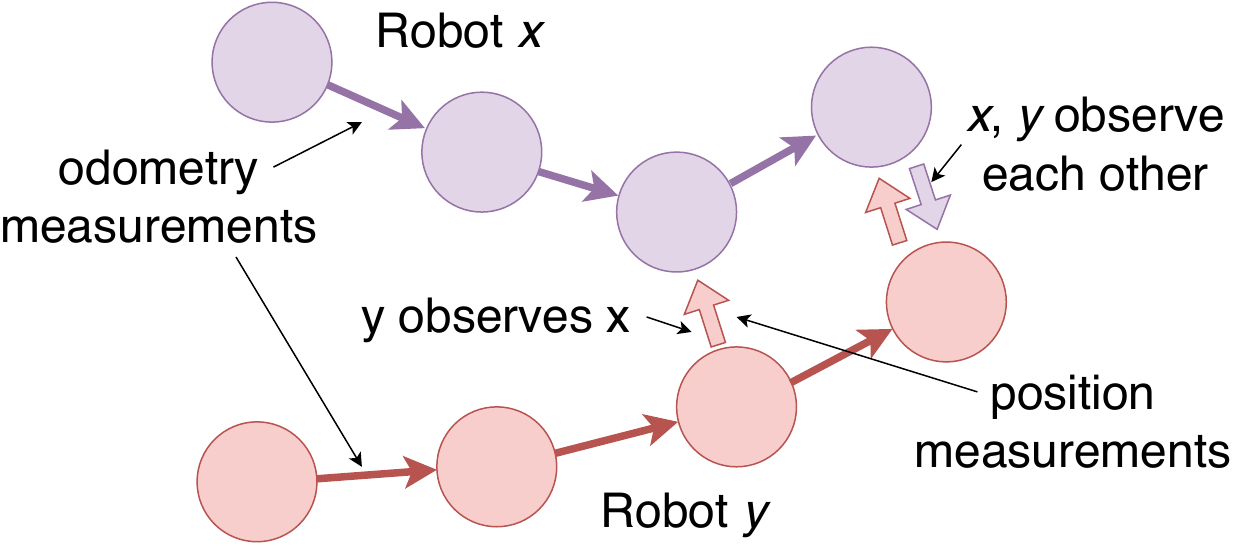}
	\includegraphics[align=c,width=0.49\linewidth]{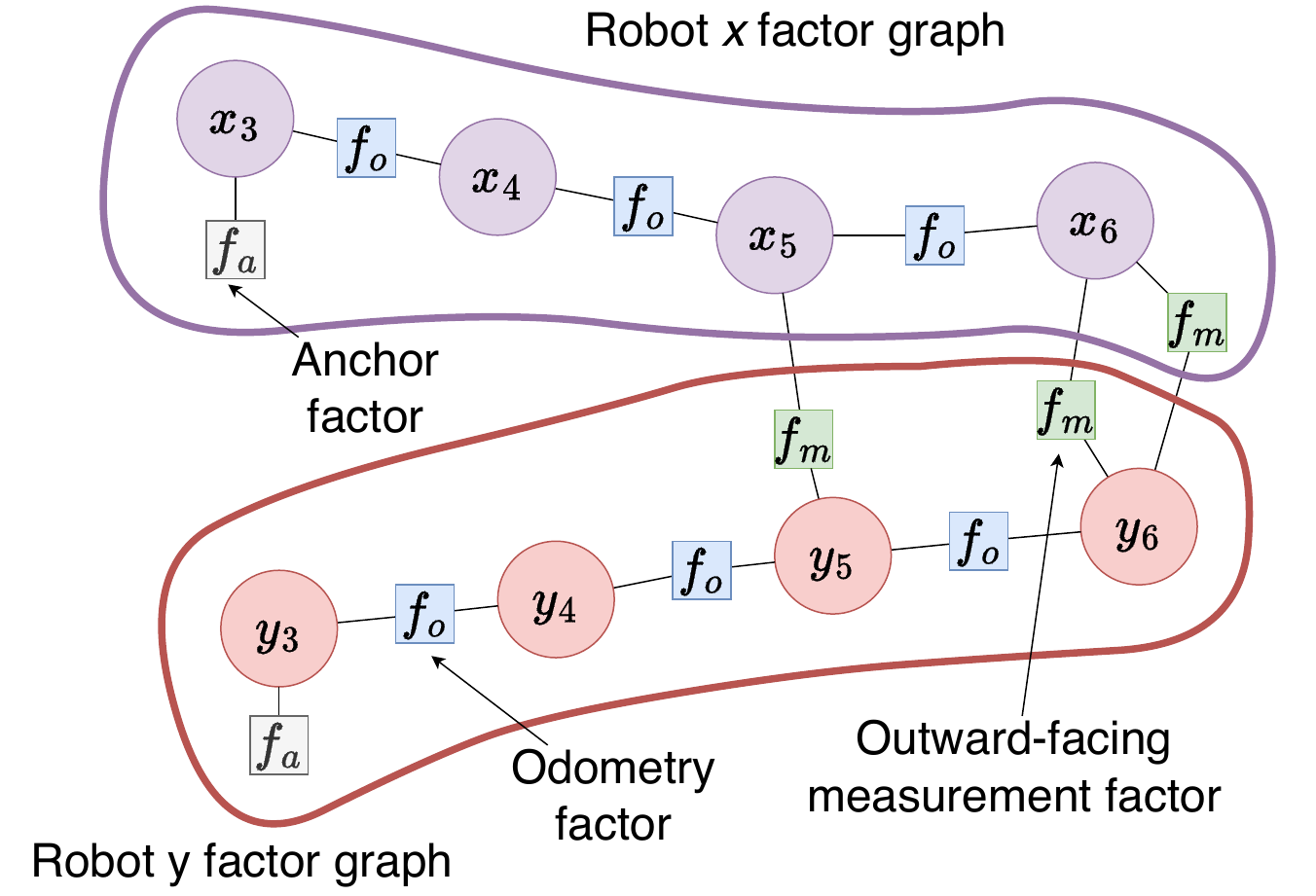}
	\caption{Robots creating connected factor graphs. Left: Two robots move on trajectories, making odometry measurements, that bring them within sensing range of each other. Right: The internal factor graphs that are created on each robot, assuming $n_{window}=4$ and some time has passed. The oldest variable node in each graph, $x_3,y_3$, has an anchor factor. Between each variable node is an odometry measurement factor node. In timestep $ts_5$, robot $y$ has observed robot $x$, creating an outward-facing relative position measurement factor. In $ts_6$, both robots have observed each other.}
	\label{fig:factorgraph}
\end{figure}

The factors linking variable nodes on different robots create the constraints that produce convergence of local pose estimates into a shared consistent state, i.e a shared reference frame.
This process is illustrated in Figure \ref{fig:factorgraph}.
Assume that robot $x$ has sensed robot $y$, this is the process that follows. It sends a list of the timesteps of outward-facing factors connected to robot $y$, which will just be $\{6\}$, for the factor connecting $x_6$ to $y_6$. Robot $y$ replies with $\{b(y_6), m_{f\to x_6}, m_{f\to x_5}\}$. Robot $x$ has now got message information needed to update the beliefs of $x_5$, $x_6$, and to create an outward message from $f_m$ attached to $x_6$.

\subsubsection{Convergence} 
\begin{figure}
	\centering
	\includegraphics[width=0.7\columnwidth]{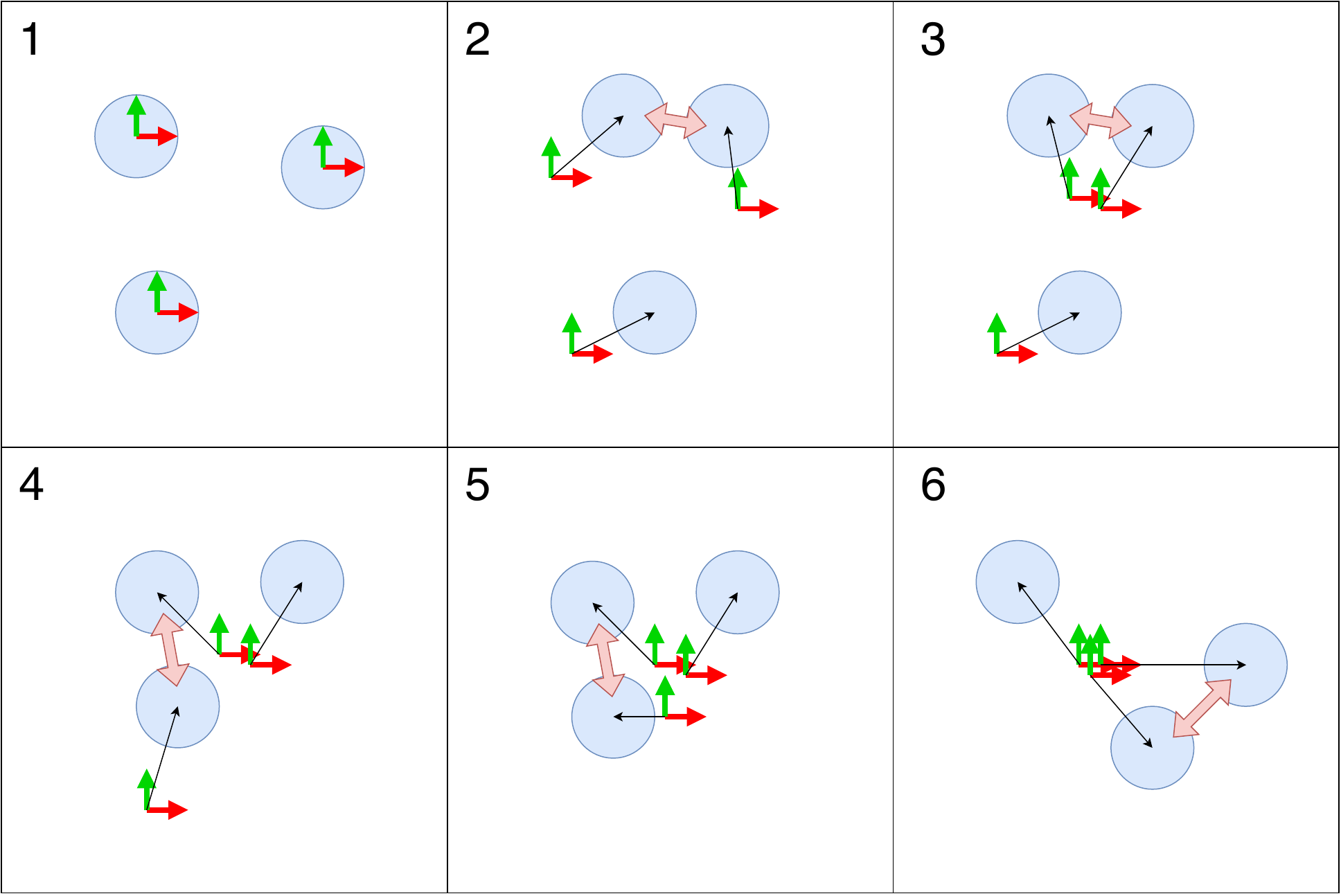}
	\caption{Illustration of shared reference frame convergence. 1) All robots start by thinking they are at the centre of the swarm. 2) Observation and communication imposes constraints on location of the swarm centroid; the top two robots communicate.. 3) .. and each robot updates its own estimate. 4) More communication imposes further constraints. 5) Origin estimates move closer.. 6) ..and approach convergence. }\label{fig:srf}
\end{figure}

As robots encounter each other and exchange messages, and individual robots perform message passing on their individual fragments of the factor graph, the pose of each robot becomes constrained against others, thus the assumed origin or reference frame of each robot converges, Figure \ref{fig:srf}. This process is dynamic, and will never completely converge, since robots are moving and have noisy perceptions, and the whole graph does not remain fully connected. The error is the mean deviation from the swarm centroid of the robot origins:
\begin{align}
	r_{error}=\frac{1}{n_{robots}}\sum_{i=1}^{n_{robots}}|\vec{r^i_{origin}}-\mu_{origin}| 
\end{align}
This is only knowable from the global perspective of the simulator. We define the convergence time $t_{conv}$ as the time taken for error to reach $r_{error}<2\sigma_{position}$ since position observation noise dominates convergence error.

To make the shared reference frame useful to a swarm system, individual agents within the swarm need to know when they can rely on estimates but they have no access to the global measure $r_{error}$. In characterising the system, we collect data on the time taken for agents to encounter at least half the robots in the swarm, denoted $t_{met\_half}$. We reason that this, with some constant, should be a reasonable proxy for the time taken to build a fully converged reference frame:
\begin{align}
	t_{convproxy} = \beta \cdot t_{met\_half}\label{eqn:beta}
\end{align}

To implement this, each robot keeps a count of the number of unique other robots it has encountered. When this exceeds half the size of the swarm, the time is noted and $t_{proxyconv}$ calculated. Only when the elapsed time is greater than this is it possible to use the shared reference frame derived $\vec{p_{robot}}$. The two example algorithms below wait for this before switching behaviour from the baseline random walk controller DSA-RW.

\subsection{Application - Shape formation}
Because each robot has access to the shared reference frame, it is easy to construct algorithms for swarm-wide shape formation. To demonstrate this, we use a simple algorithm called DSA-SF (Distributed Spatial Awareness - Shape Formation) where the shape is defined functionally, $f_{in\_shape}(p_{robot})$, and shown in Algorithm \ref{alg:shape}. Each robot has two modes of behaviour; when not inside the shape according to its estimated position $\vec{p_{robot}}$, it uses the baseline behaviour DSA-RW. When inside the shape, it slows down to $v_{slow}=0.1\cdot v_{fast}$, and if there are any neighbours, it moves towards them, causing classic swarm aggregation. 
\begin{algorithm}\caption{DSA-SF Shape formation}\label{alg:shape}
\normalsize
\begin{algorithmic}
\If {not \Call{in\_shape}{$p_{robot}$}}
	\State $\vec{v_{cmd\_vel}} =  v_{fast}\cdot\text{DSA-RW}_{direction}$
\Else 
	\If {$n_{neighbours} > 0$}
		\State $\vec{v_{cmd\_vel}} = (v_{slow},\angle_{neighbours})$
	\Else
		\State $\vec{v_{cmd\_vel}} =  v_{slow}\cdot\text{DSA-RW}_{direction}$
	\EndIf
\EndIf
\end{algorithmic}
\end{algorithm}
We define some simple shapes such as a circle, vertical and horizontal lines, and wavy lines, and switch between them at intervals.

\subsection{Application - Intralogistics}
As noted above, even random walkers are capable of performing simple swarm logistics operations. We want to enhance the performance of such systems using the shared reference frame. A key component to a logistics system is knowledge of the location of cargo carriers. Carriers $C$ are dynamic objects that may move. Each robot maintains a set of Gaussian variables and an associated time of observation $c(i)\equiv(\vec{x_{carrier}^{i}},t_{observed}^i), i \in C$, one for each possible carrier. When a robot observes a carrier $k$, it sets the tuple $c(k)=((\vec{p_{robot}}+\vec{p_{carrier}},\sigma^2_{position}+\sigma^2_{robot})_G,t)$. Each time a robot exchanges messages with another robot, it sends  a list of the observation times $t_{observed}^i,i\in C$ it has, including $t=0$ for carriers for which it has no observation. The other robot replies with any carrier observations it has that are more recent, these are used to replace the older observations. More recent observations are privileged, even if they may have greater positional uncertainty, since the carrier may have moved.
\begin{algorithm}\caption{DSA-KE Knowledge Enhancement}\label{alg:dsake}
\normalsize
\begin{algorithmic}
\If {any $t_{observed}^i = 0, i \in C $}
	\State $\vec{v_{cmd\_vel}} =  v_{fast}\cdot\text{DSA-RW}_{direction}$
\Else 
	\State $j=argmin(t_{observed}^i,i\in C)$
	\State $\vec{d}=\vec{x_{carrier}^j}-\vec{p_{robot}}$
	\State $\vec{v_{cmd\_vel}} =  v_{fast}\cdot\vec{d}$
\EndIf
\end{algorithmic}
\end{algorithm}

The way the robots move has an impact on the acquisition of knowledge about the environment. When the swarm has no knowledge, the goal is to cover as much of the arena area as possible. For this, we use the baseline behaviour DSA-RW.
Once the swarm has a certain level of awareness, it becomes possible to use this information to guide swarm exploration so as to maximise knowledge. This behaviour is called DSA-KE (Distributed Spatial Awareness - Knowledge Enhancement), and is identical to DSA-RW until a robot has some information about all the carriers in the environment. At that point, instead of choosing a direction at random, a robot will choose the direction of the carrier which it has oldest knowledge about, thus the swarm as a whole seeks to minimise overall uncertainty, Algorithm \ref{alg:dsake}.

In order to test the quality of knowledge acquisition in a dynamic environment, we make the carriers move position. This is specified with a single parameter, the aggregate mean carrier velocity $v_{c\_agg}$. Given $n_{carrier}$ carriers, a carrier is selected at random and moved a fixed distance of \SI{1}{m} into a random location at a velocity of $v_{c\_agg}\cdot n_{carrier}$. The mean perception error of the swarm is given by:
\begin{align}
s_{error} &= \frac{1}{n_{robots}n_{carrier}}\sum_{j=1}^{n_{robots}}\sum_{i=1}^{n_{carrier}}|c(i)_{est}^j-c(i)_{gt}|
\end{align}
where $c(i)_{gt}$ is the ground truth position of a carrier, transformed into the swarm frame.

\section{Results}\label{sec:results}
We measured the performance of various metrics of interest. In each case, simulations were run 50 times with different random seeds for each datapoint. Standard parameters are shown in Table \ref{tab:params}. 
\begin{table}[h]\centering\caption{Simulation parameters}\label{tab:params}
\begin{tabulary}{\columnwidth}{lll}
	\hline
	Parameter & Value & Description \\
	\hline
	$n_{window}$ & $10$ & Max number of variable nodes in local factor graph \\
	$t_{node}$ & \SI{0.5}{s} & New variable node creation interval\\
	$r_{sense}$ & \SI{0.5}{m}& Object and robot sense and communication radius \\
	$\sigma_{velocity}$ & \SI{0.1}{m/m} & Velocity sense noise (metres per metre travelled)\\
	$\sigma_{position}$ & \SI{0.02}{m} & Position sense noise\\
	$v_{fast}$ & \SI{0.5}{ms^{-1}} & Movement performing DSA-RW and DSA-KE\\
	$v_{slow}$ & \SI{0.05}{ms^{-1}} & Movement within shape while performing DSA-SF \\
	\hline
\end{tabulary}
\end{table}
\subsection{Convergence, Computation, and Bandwidth}
We examine how long the distributed factor graph takes to converge, $t_{conv}$, and how much computation and bandwidth is used in different scenarios. An example video of 10 robots reaching convergence is available at  \url{https://youtu.be/wo5tLKpMmes}.

\begin{figure}
	\centering
	\includegraphics[trim={12 0 10 13},clip,width=0.482\linewidth]{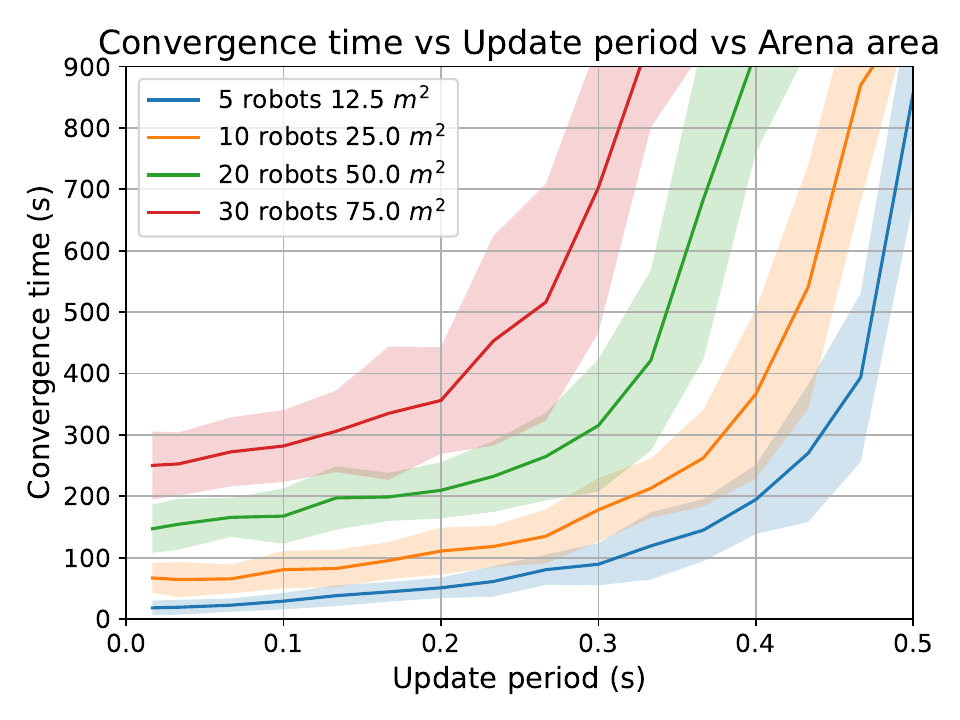}
	\includegraphics[trim={15 5 7 12},clip,width=0.49\linewidth]{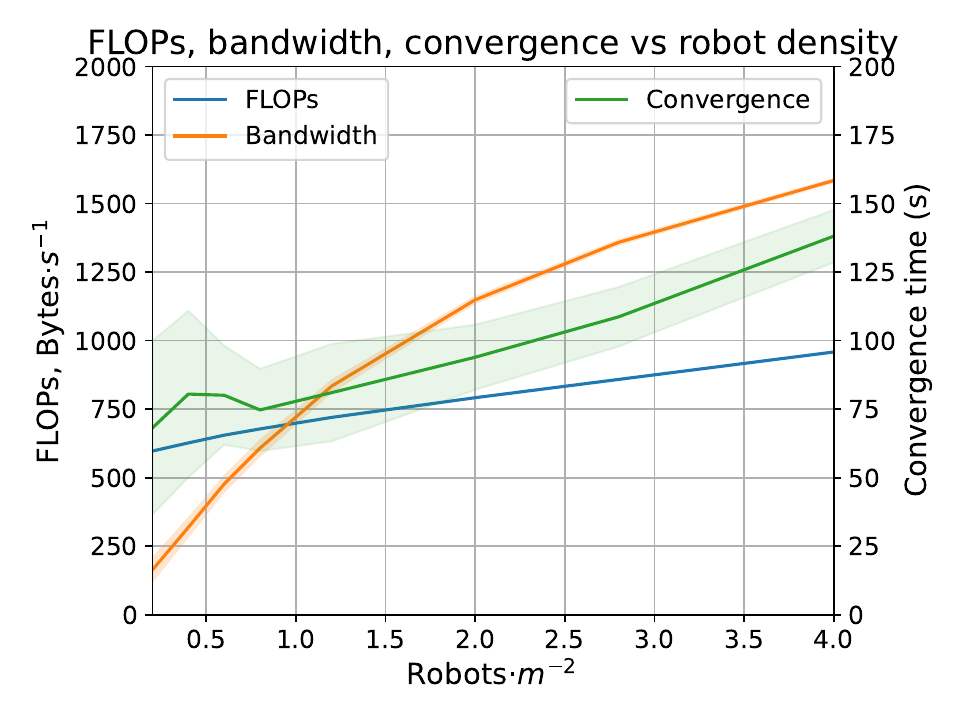}
\caption{Left: Convergence time with different factor graph update rates and different arena areas, with fixed robot density. Reduction in convergence time is minimal below \SI{0.1}{s} update period. Right: For a fixed update period of \SI{0.1}{s} and fixed arena area of \SI{25}{m^2}, computation, convergence time, and particularly bandwidth are dependent on robot density.}
\label{fig:conv}
\end{figure}

We look at different $t_{message}$ update periods of the factor graph, and different numbers of robots in an arena sized to keep a constant robot density of \SI{0.4}{m^{-2}}. See Figure \ref{fig:conv}. Convergence takes longer when there are robots over a larger area, this is expected, since all robots have to be able to influence each other, though not necessarily by direct communication. Density has little effect on convergence, the dominating factors are update rate and arena area. There is little gain from rapid updates over \SI{10}{Hz}.

Computation and bandwidth are both proportional to update frequency so we use a fixed sized arena of \SI{25}{m^2} and update rate of \SI{10}{Hz} to look at other factors. All operations are performed in 32 bit floating point. Since we are working in 2D space with linear factors, the precision matrix $\Lambda$ can be represented as a single number. Belief update needs 3 operations per attached factor (Eqn \ref{eqn:bel}). Factor message generation only needs computation for measurement factors; $2\text{(Eqn \ref{eqn:alpha})}+1+2+2+1\text{(Eqn \ref{eqn:m1})}+2+2+1\text{(Eqn \ref{eqn:m2})}=13$ operations. Assuming a naive message protocol, every request has an overhead of 12 bytes, and 4 bytes per outward-facing factor, and each response has an overhead of 12 bytes, and 16 bytes per returning belief and factor-to-variable message. 

 There is a weak dependence of computation on density, as there are more chances to create additional outward-facing factors. Bandwidth is strongly dependent on robot density, as there are many more opportunities for a robot to communicate.
It should be noted that the raw figures for supporting the shared reference frame are remarkably low; for a \SI{25}{m^2} arena, the swarm will converge in less than \SI{60}{s}, with each robot using only a few hundred floating point operations and exchanged message bytes per second. This is achievable even on low cost processors.

\begin{figure}
	\centering
	\includegraphics[trim={14 0 12 16},clip,width=0.492\linewidth]{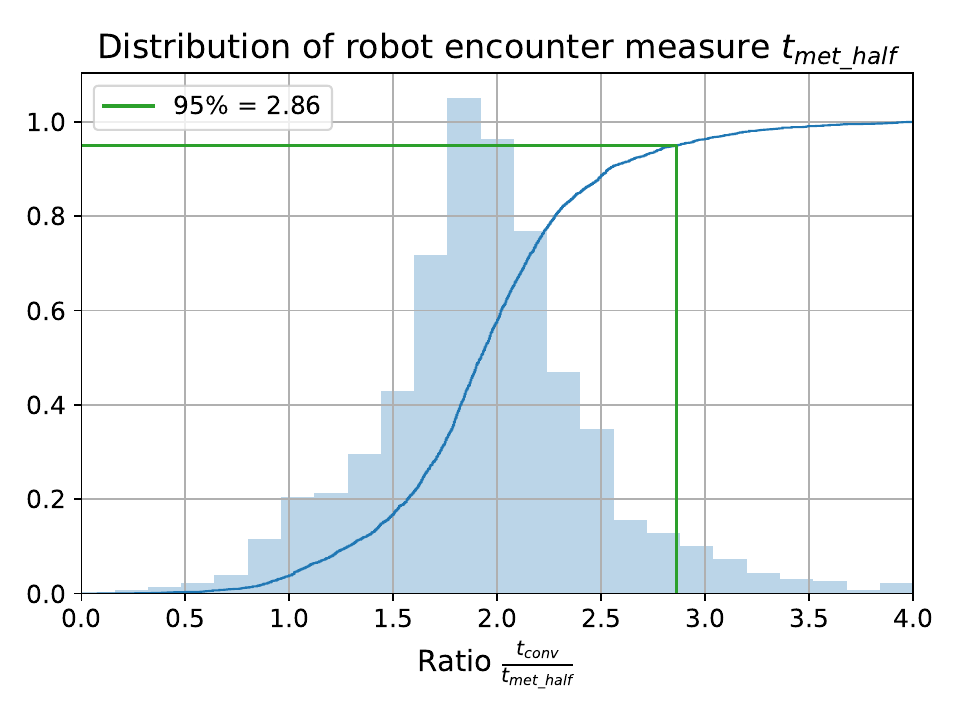}
	\includegraphics[trim={16 0 0 13},clip,width=0.495\linewidth]{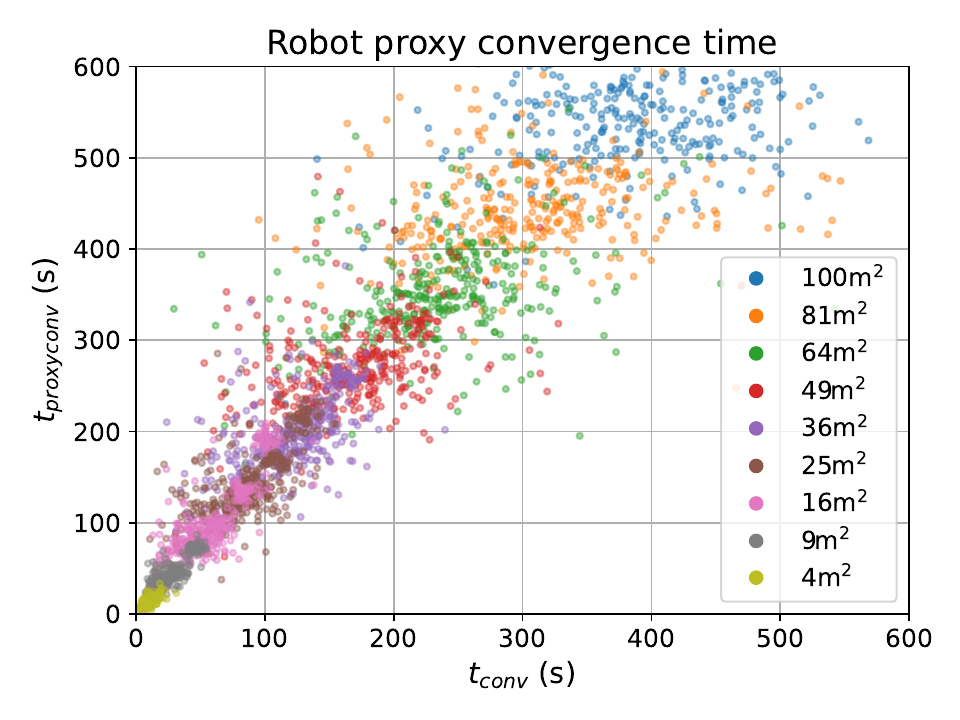}
\caption{The distribution of the robot encounter measure $t_{met\_half}$ over 2886 simulations with different numbers of robots between 5 and 100, and different arena sizes between \SI{4}{m^2} and \SI{100}{m^2}. Using a constant $\beta=3$ ensures the proxy convergence time $t_{proxyconv}$ exceeds the true convergence time $t_{conv}$ in more than 95\% of simulation. Right scatter plot coloured according to arena area.}
\label{fig:char}
\end{figure}

In order to determine an appropriate value for $\beta$ (Eqn \ref{eqn:beta}), we ran a set simulations over different numbers of  robots between 5 and 100, and different arena sizes between \SI{4}{m^2} and \SI{100}{m^2}, fixing the update period $t_{message}=0.1$\,s. Simulations were run for 1000 simulated seconds. Out of 4200 simulations, 2886 were able to find an initial placement for the robots in the arena size, and reached convergence within the simulation run time. We can see from the graphs in Figure \ref{fig:char} that using a value of $\beta=3$ ensures the proxy measure $t_{proxyconv}$ exceeds the true measure $t_{conv}$ in 95\% of simulations. 

\subsection{Shape formation}
\begin{figure}
	\centering
	\includegraphics[width=0.16\linewidth]{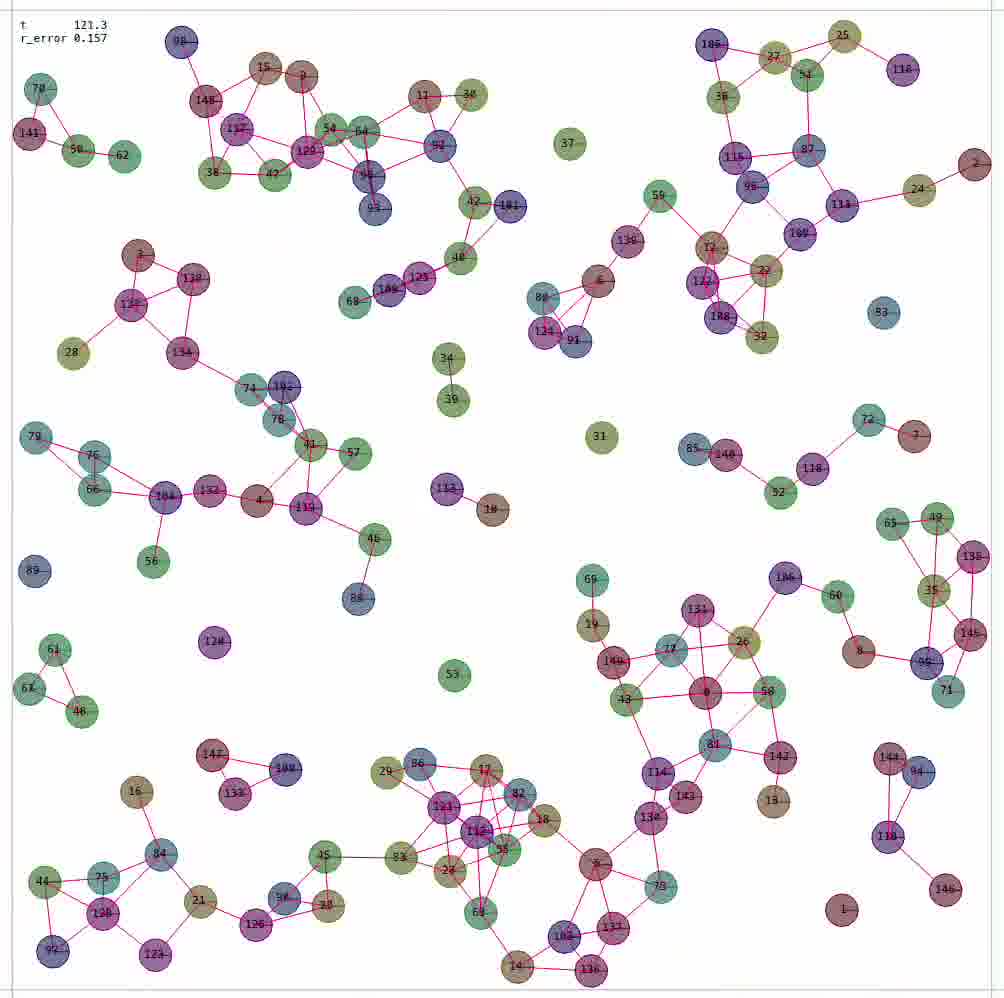}
	\includegraphics[width=0.16\linewidth]{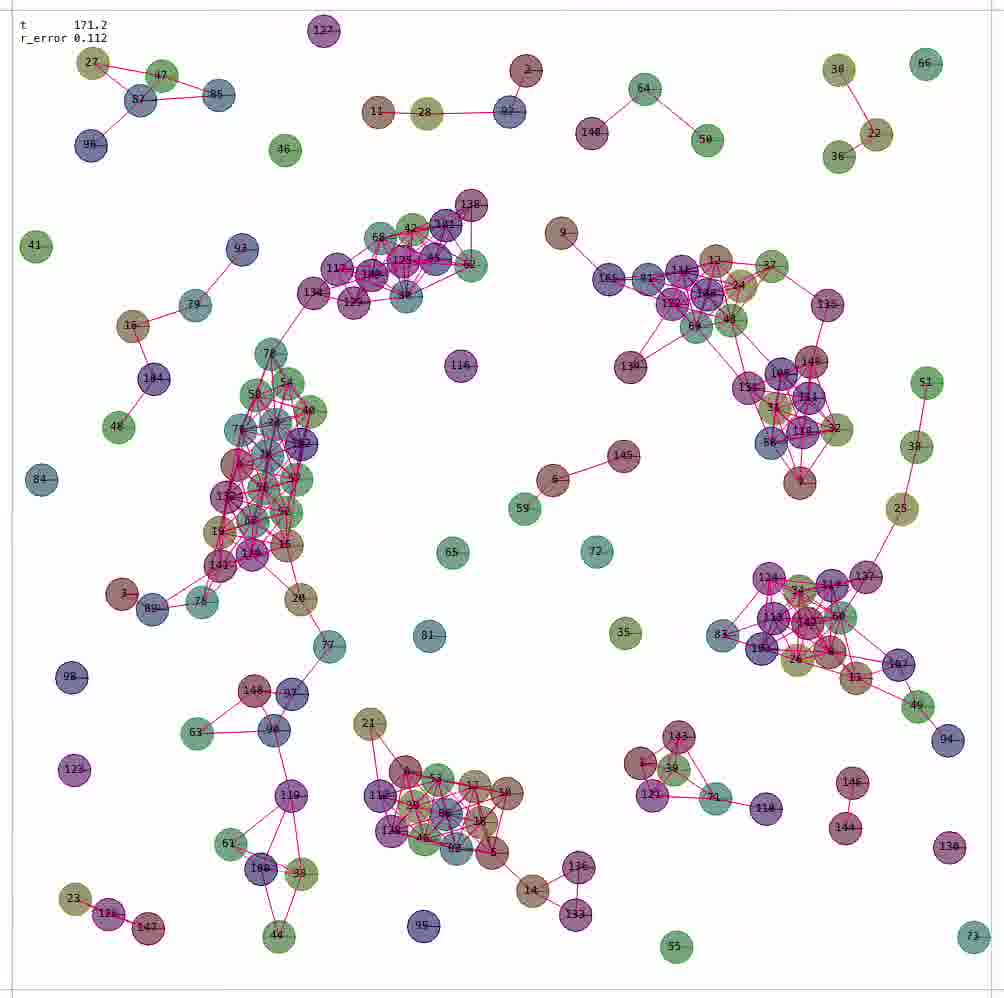}
	\includegraphics[width=0.16\linewidth]{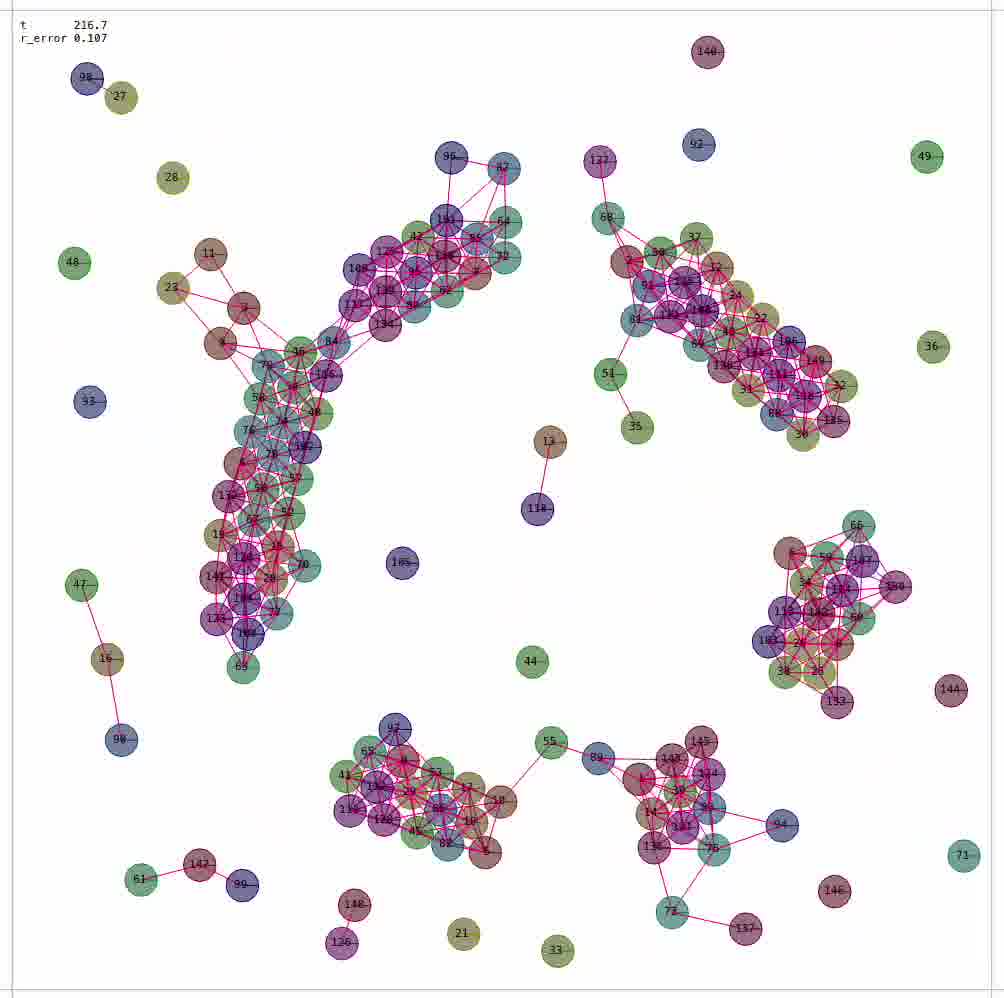}
	\includegraphics[width=0.16\linewidth]{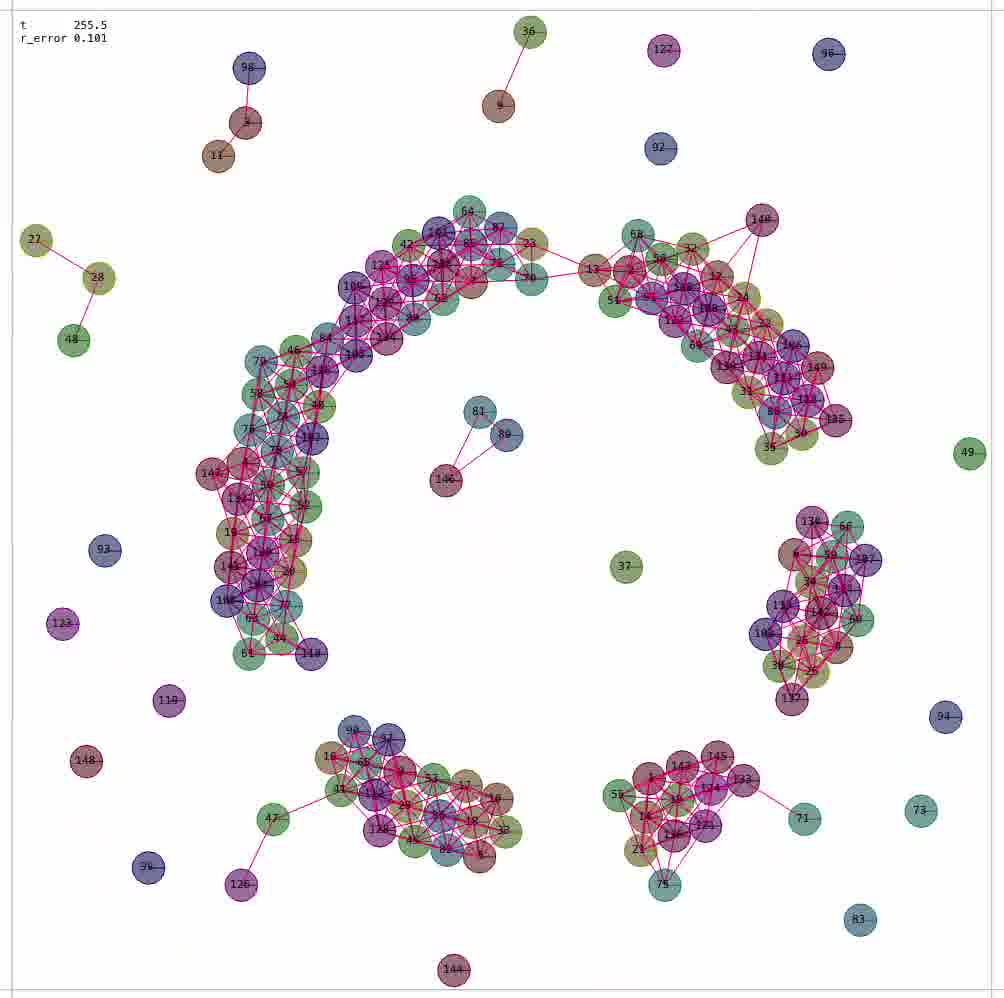}
	\includegraphics[width=0.16\linewidth]{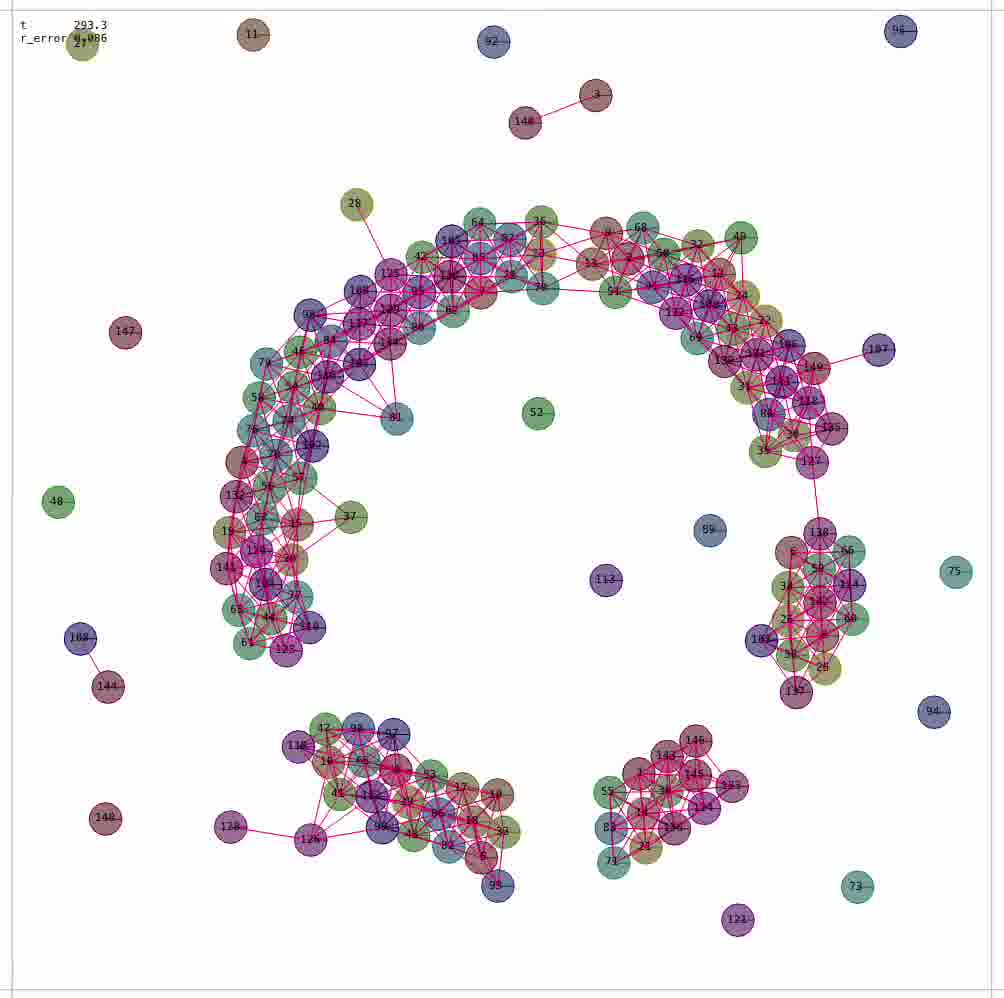}
	\includegraphics[width=0.16\linewidth]{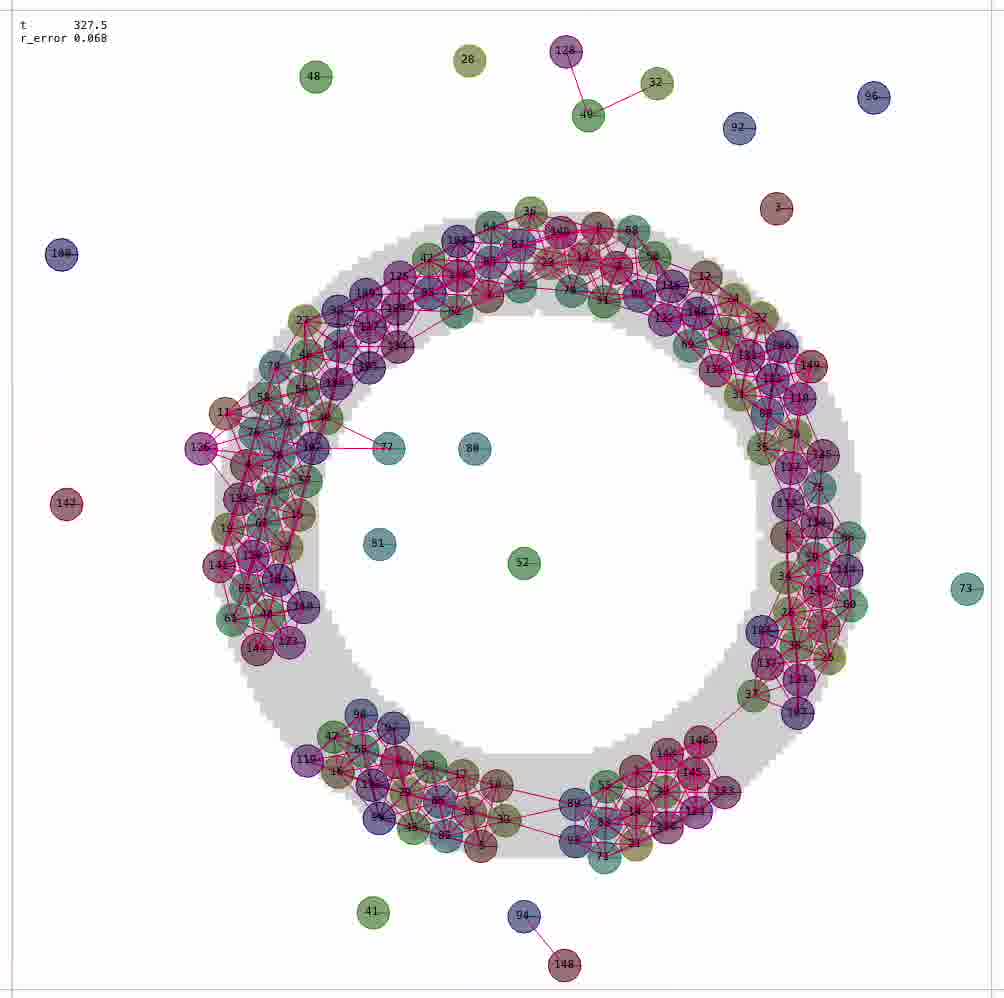} \\
	\includegraphics[width=0.16\linewidth]{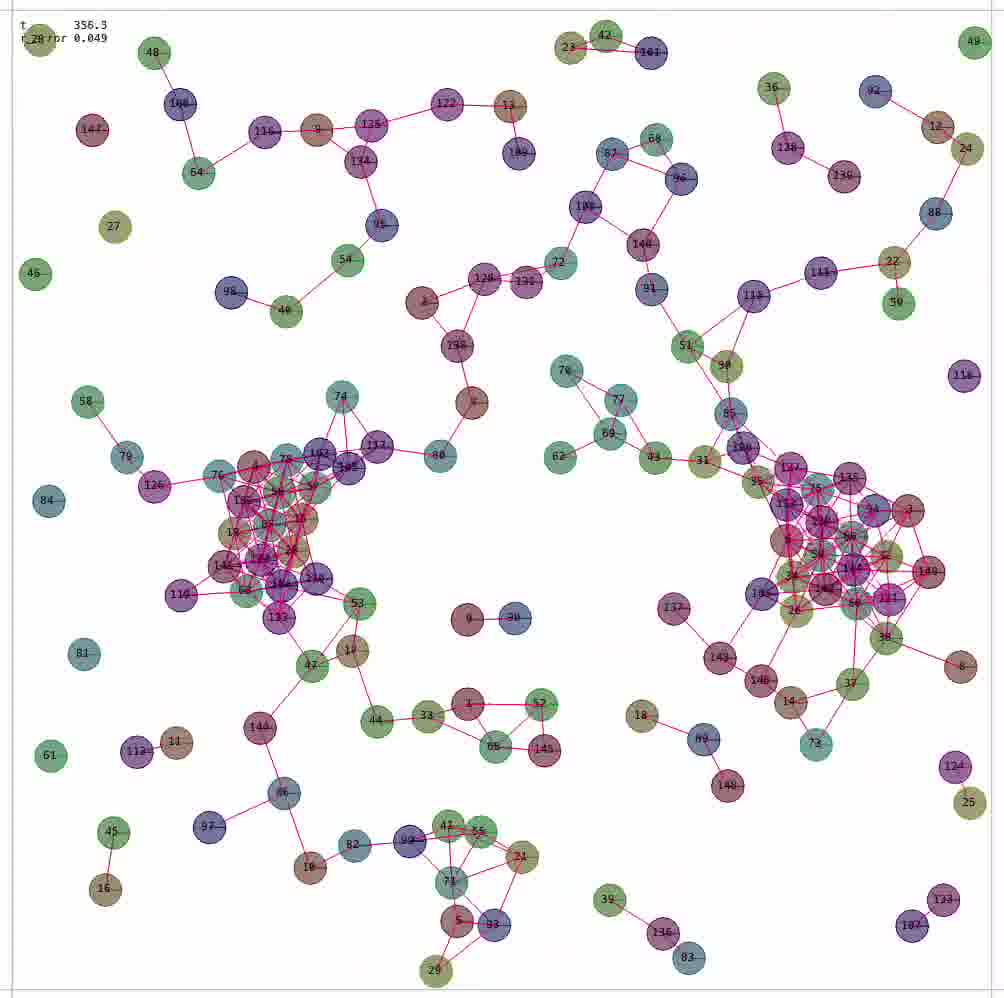}
	\includegraphics[width=0.16\linewidth]{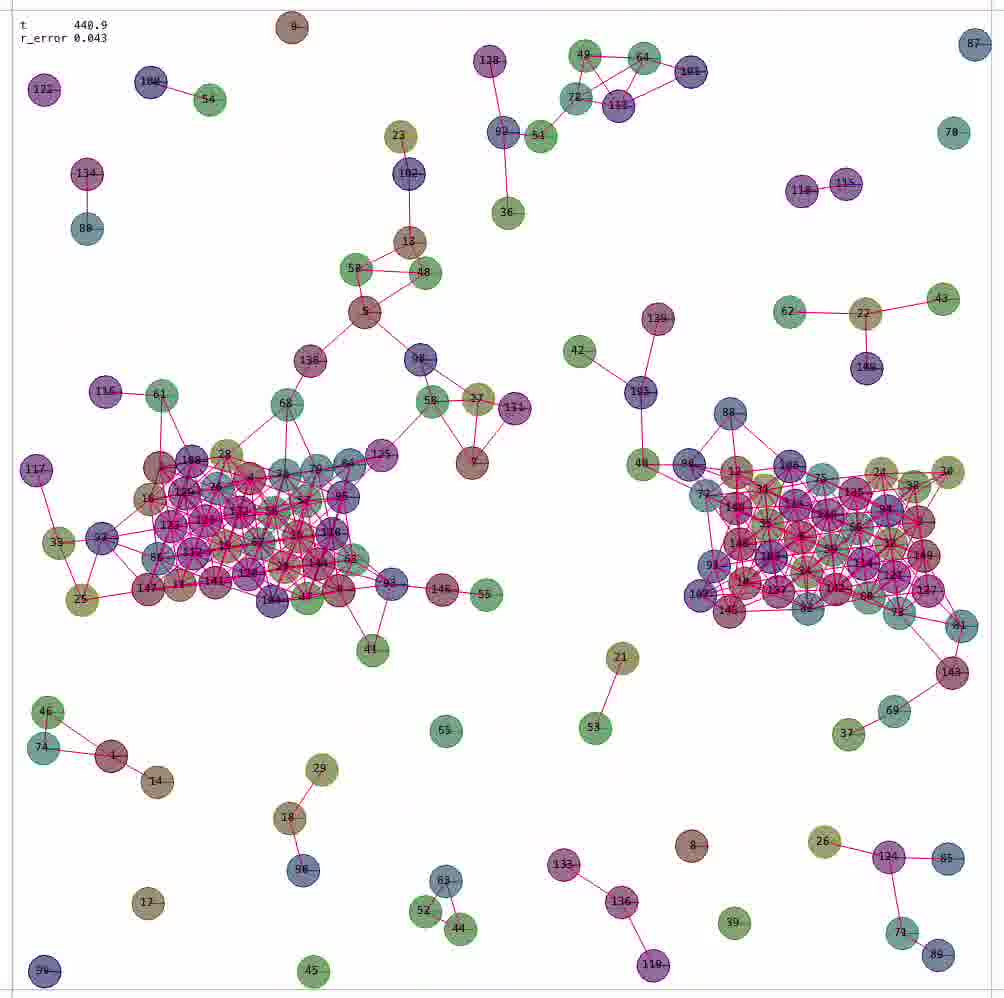}
	\includegraphics[width=0.16\linewidth]{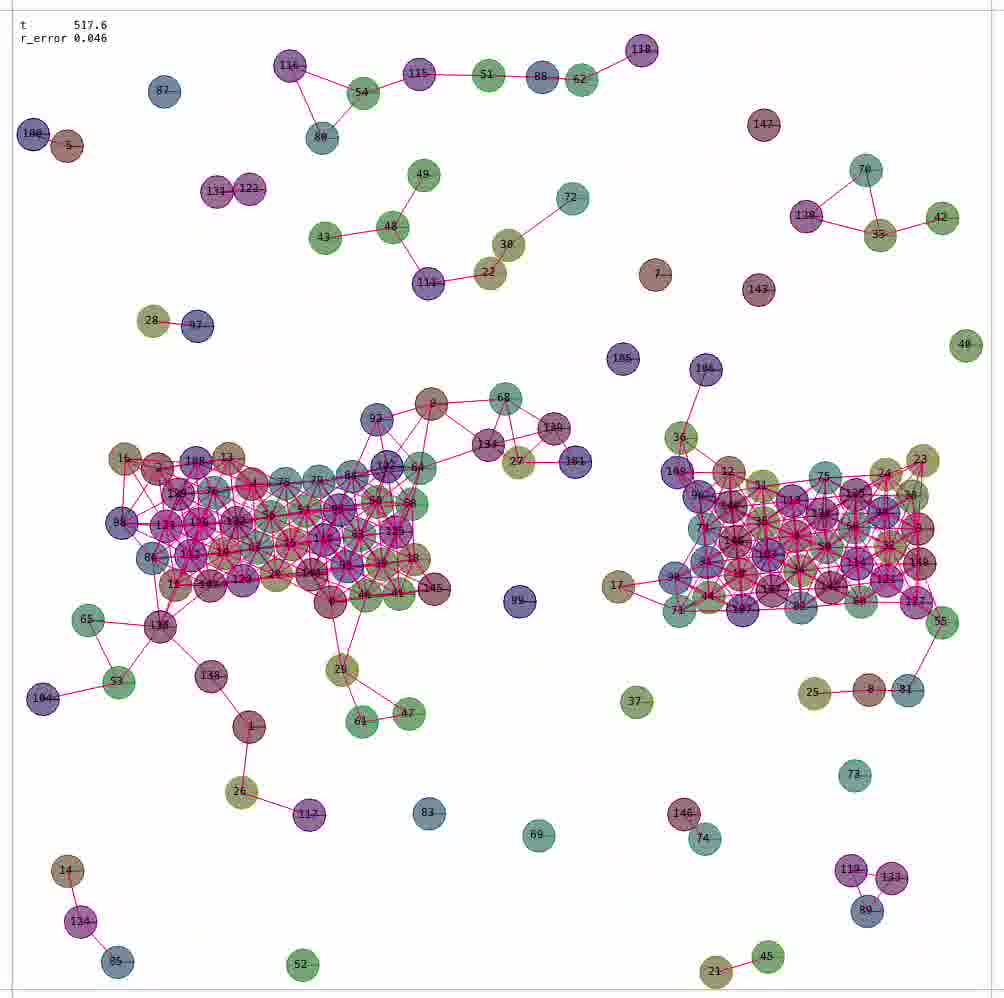}
	\includegraphics[width=0.16\linewidth]{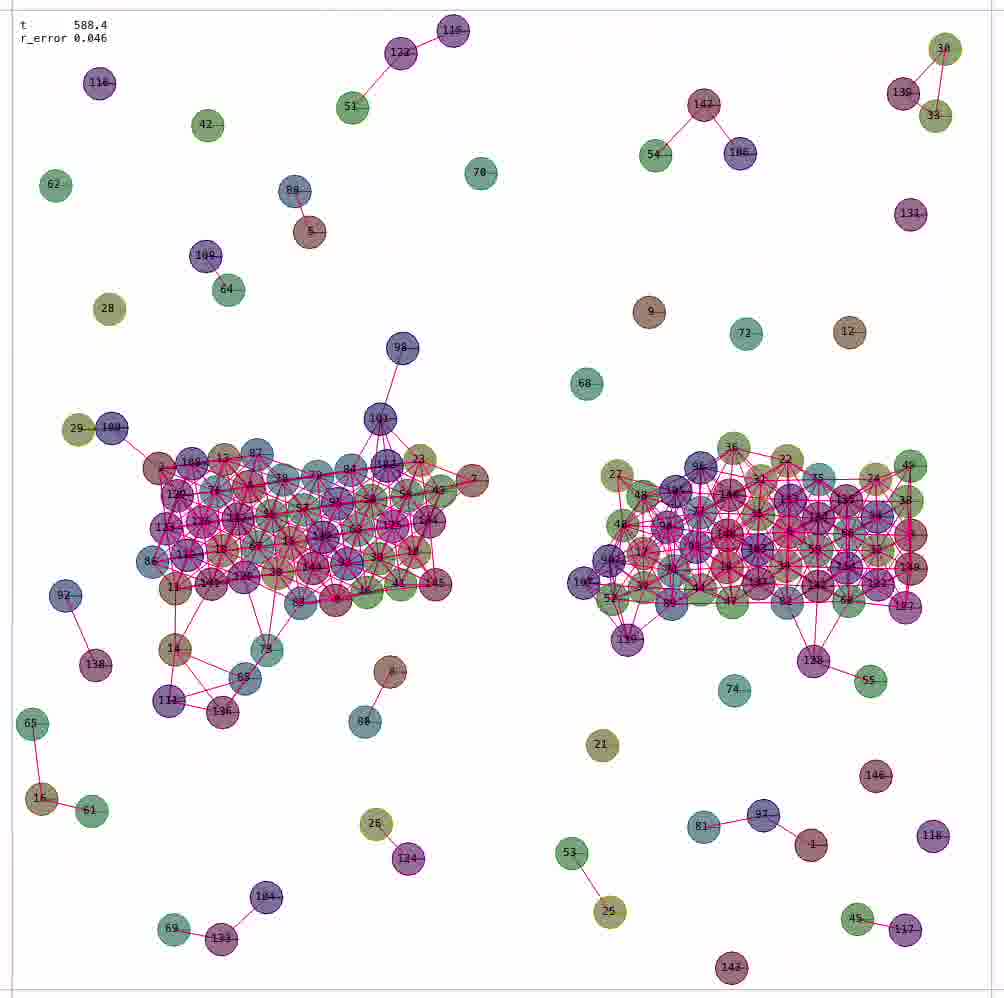}
	\includegraphics[width=0.16\linewidth]{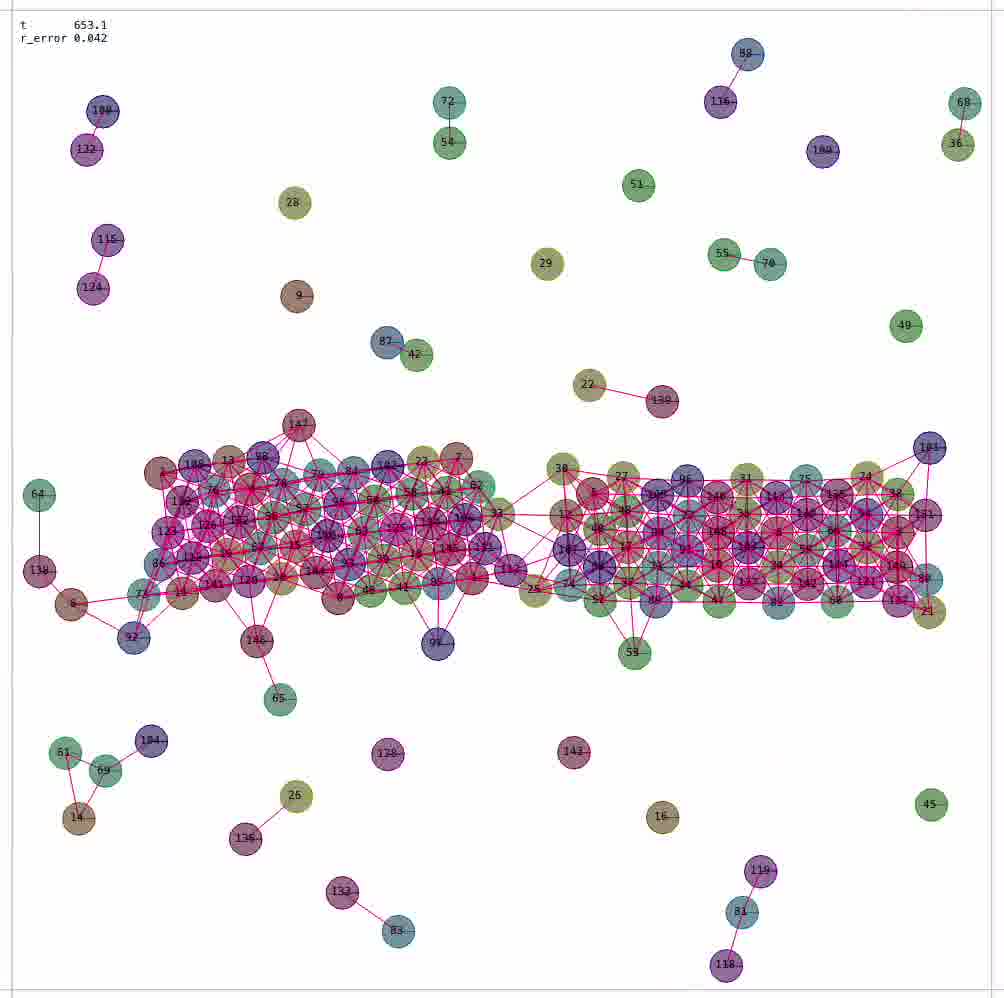}
	\includegraphics[width=0.16\linewidth]{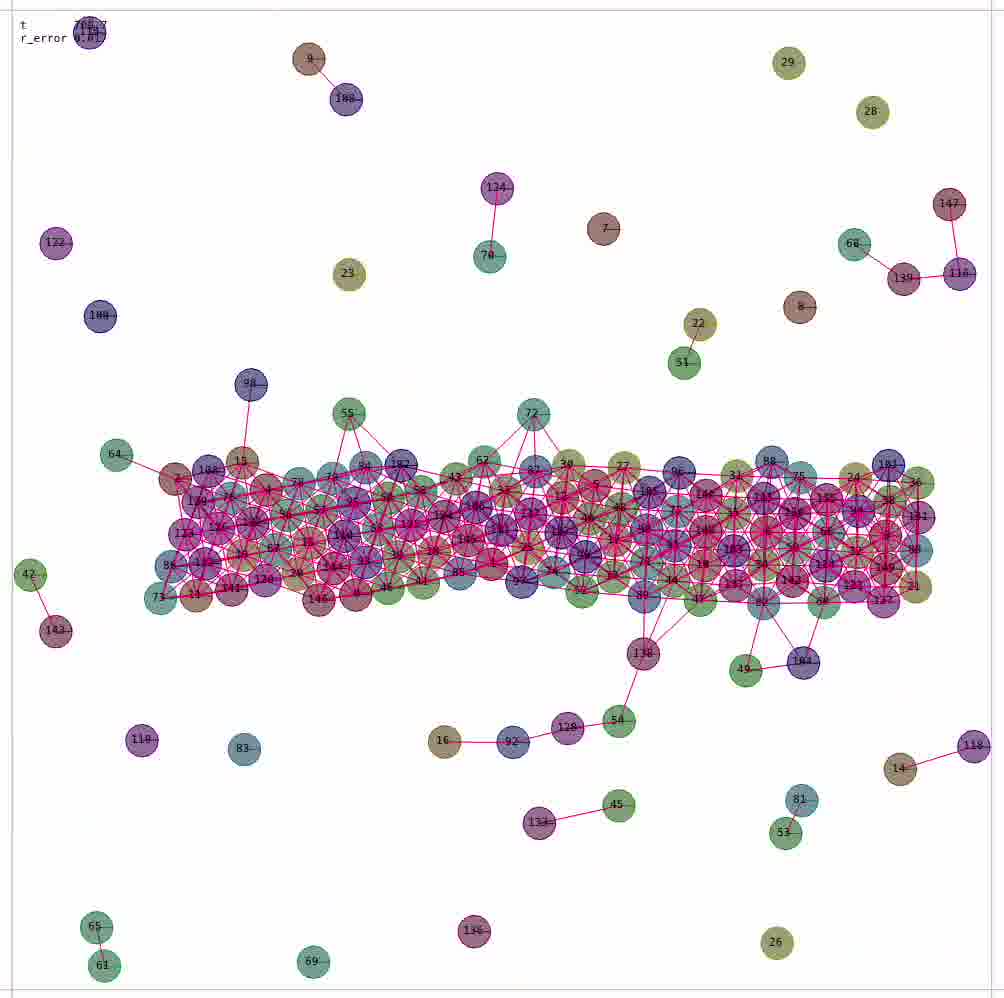}\\
	\includegraphics[width=0.16\linewidth]{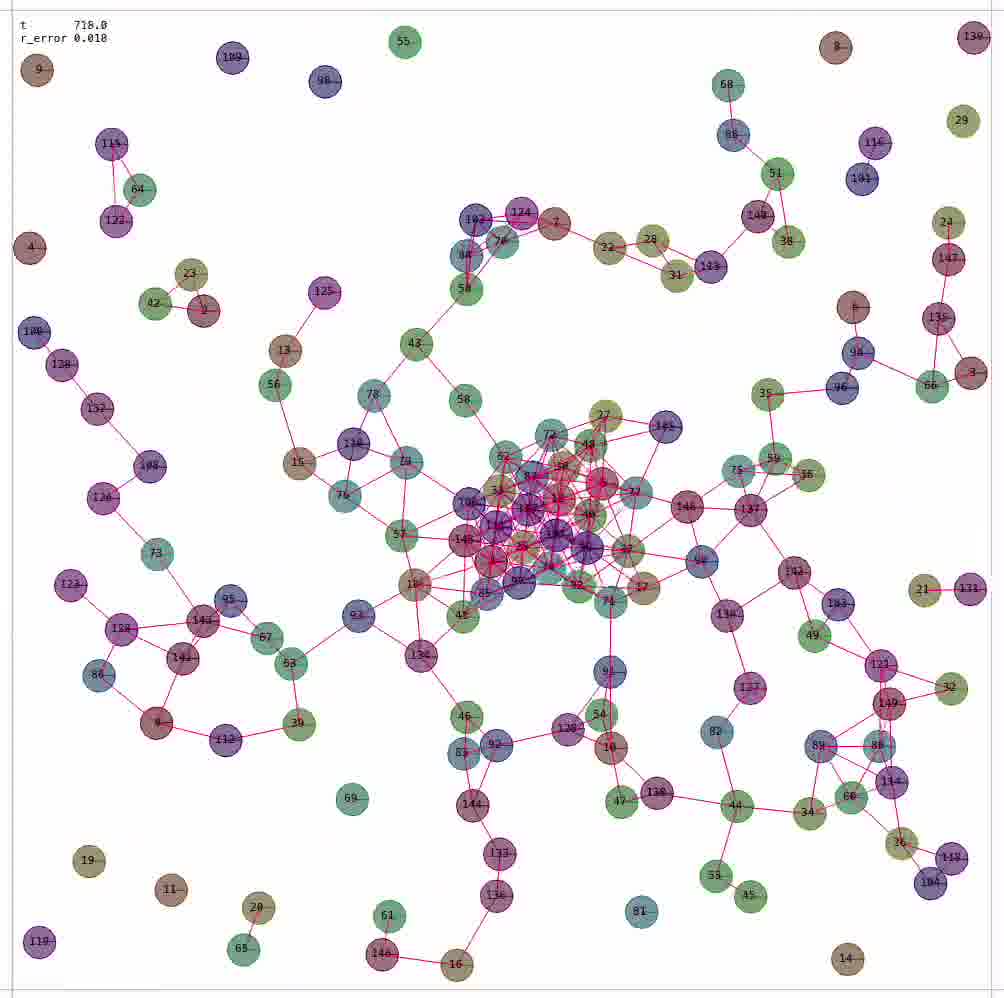}
	\includegraphics[width=0.16\linewidth]{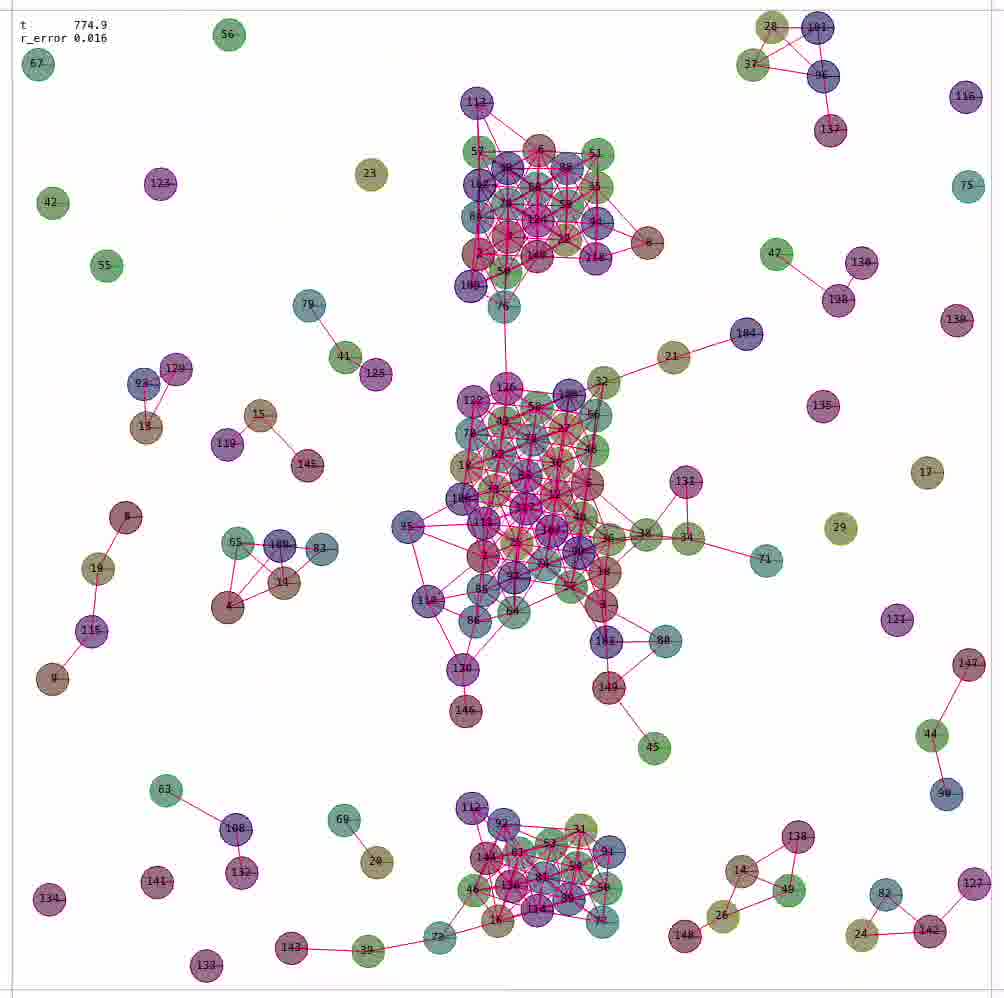}
	\includegraphics[width=0.16\linewidth]{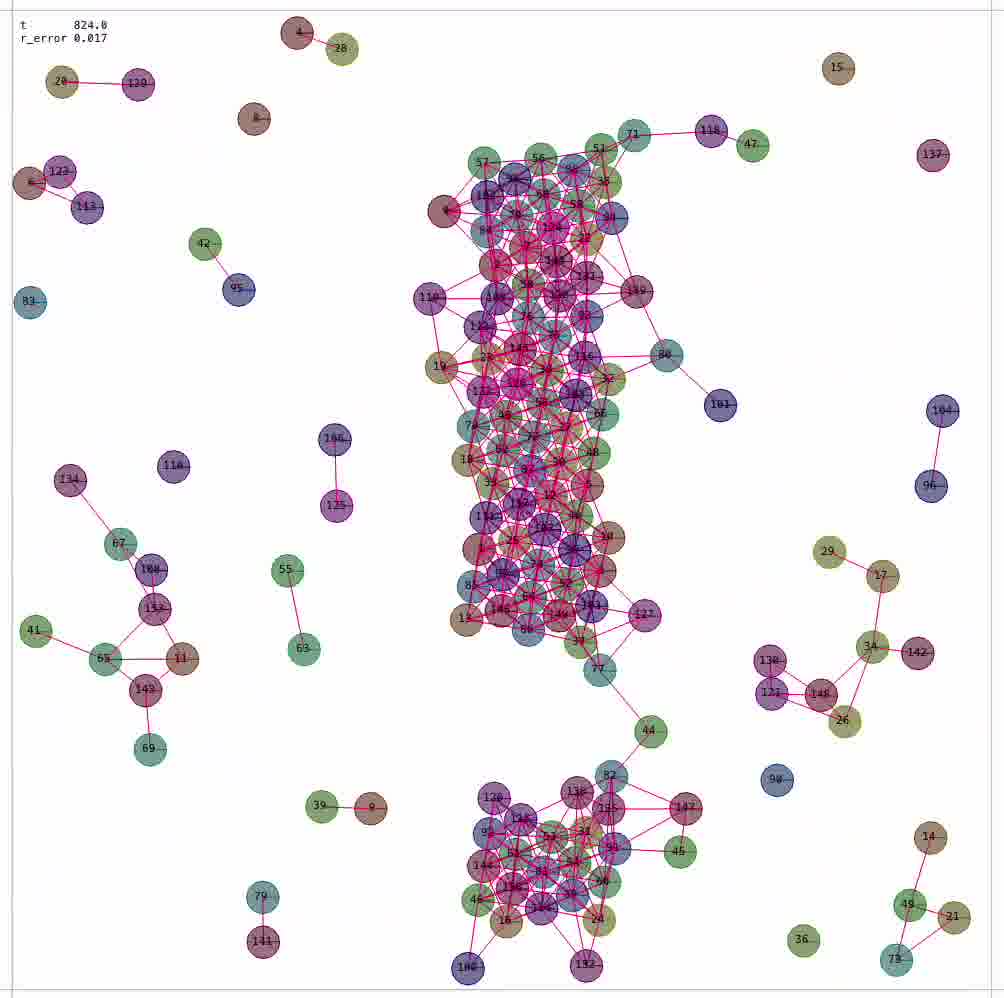}
	\includegraphics[width=0.16\linewidth]{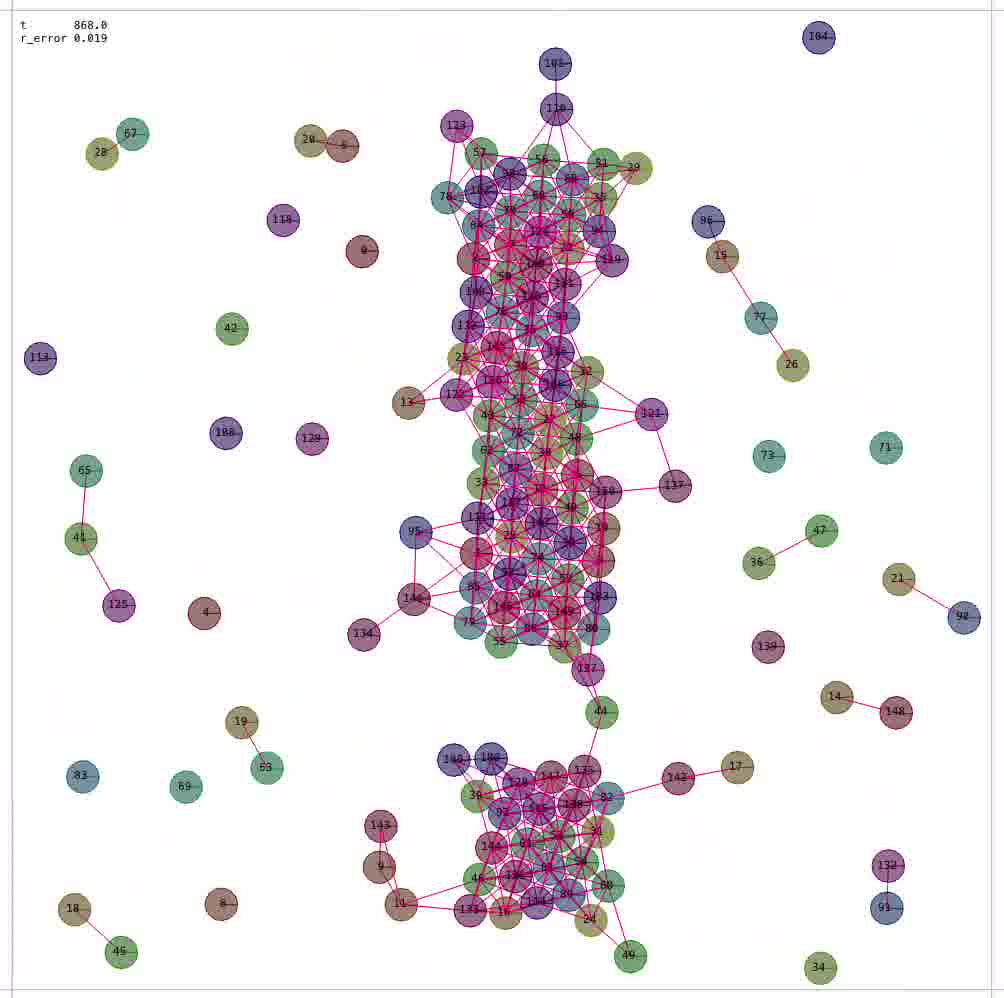}
	\includegraphics[width=0.16\linewidth]{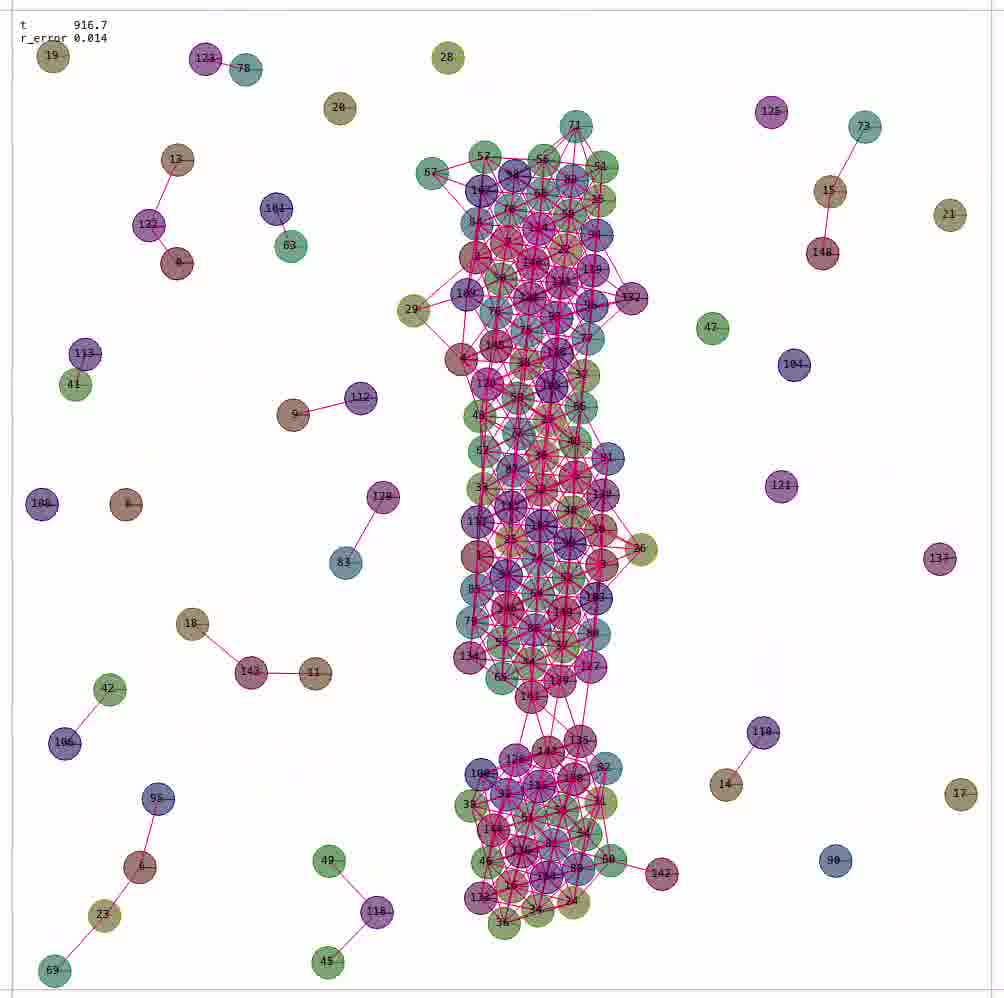}
	\includegraphics[width=0.16\linewidth]{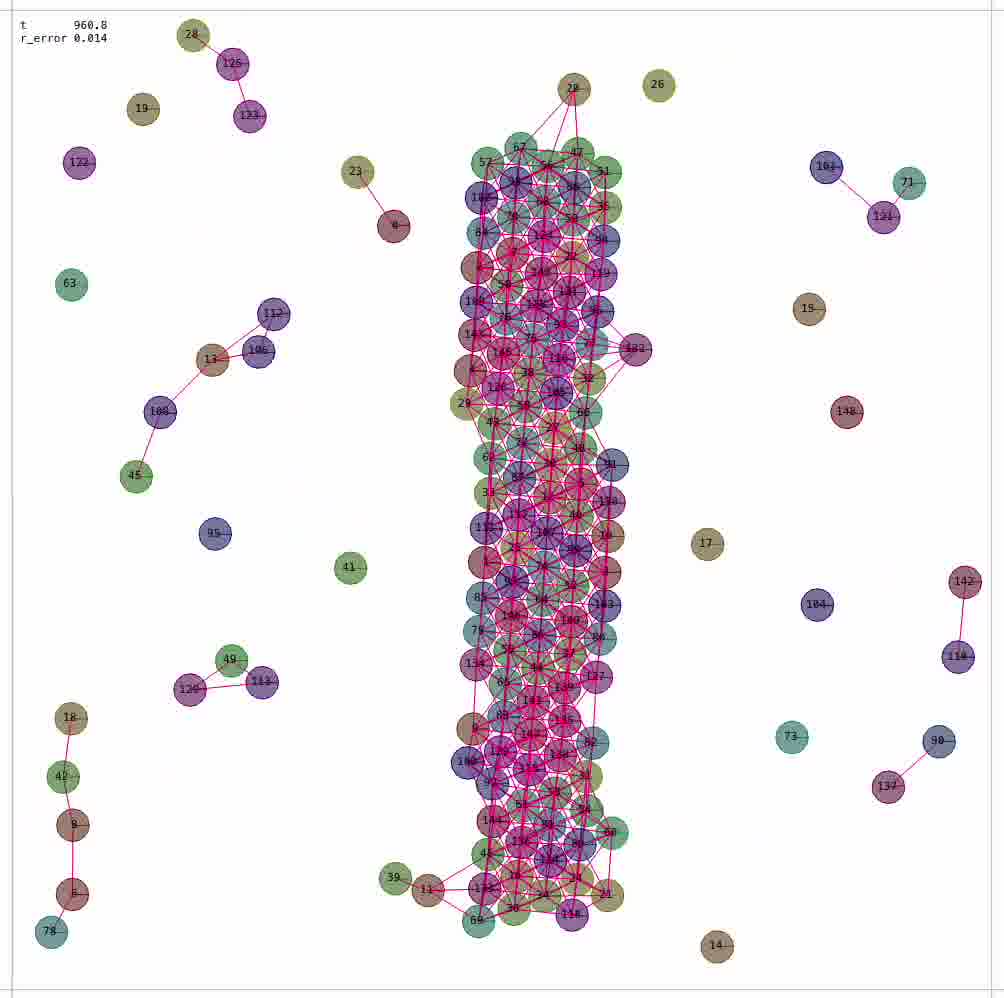}\\
	\includegraphics[width=0.16\linewidth]{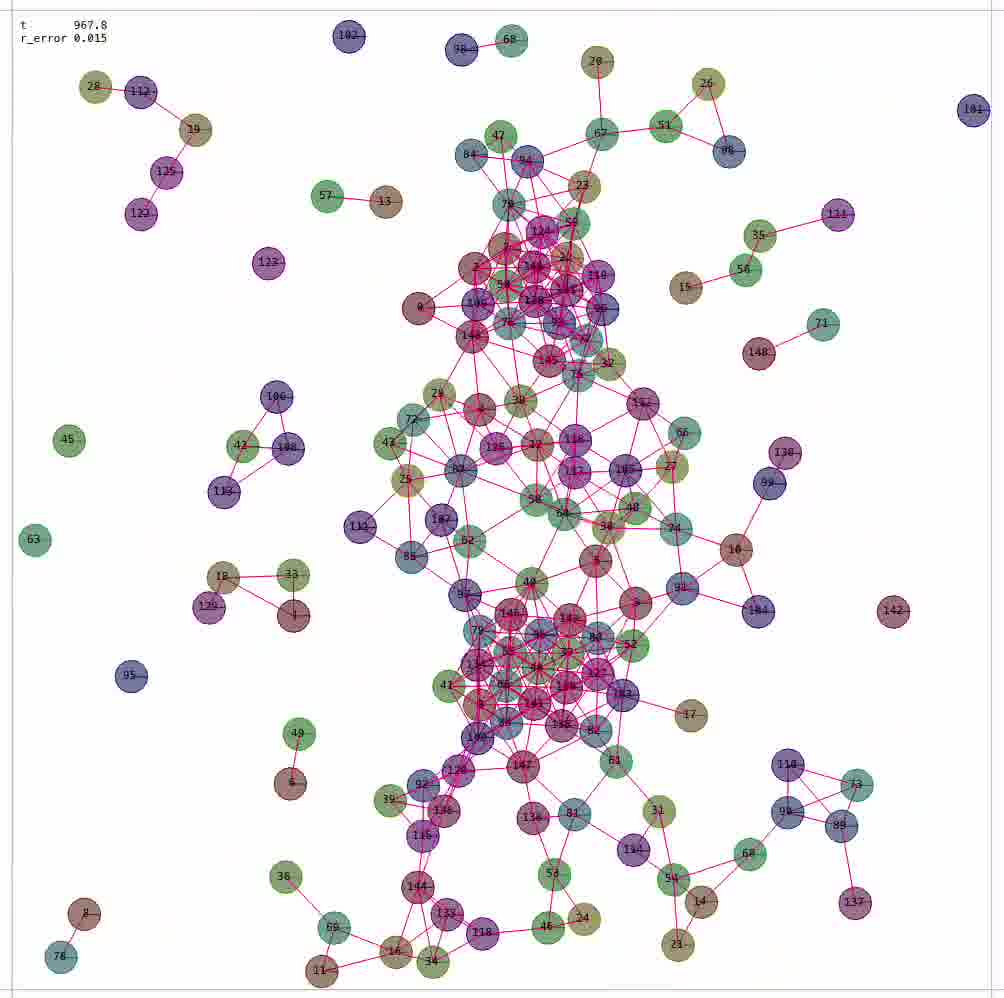}
	\includegraphics[width=0.16\linewidth]{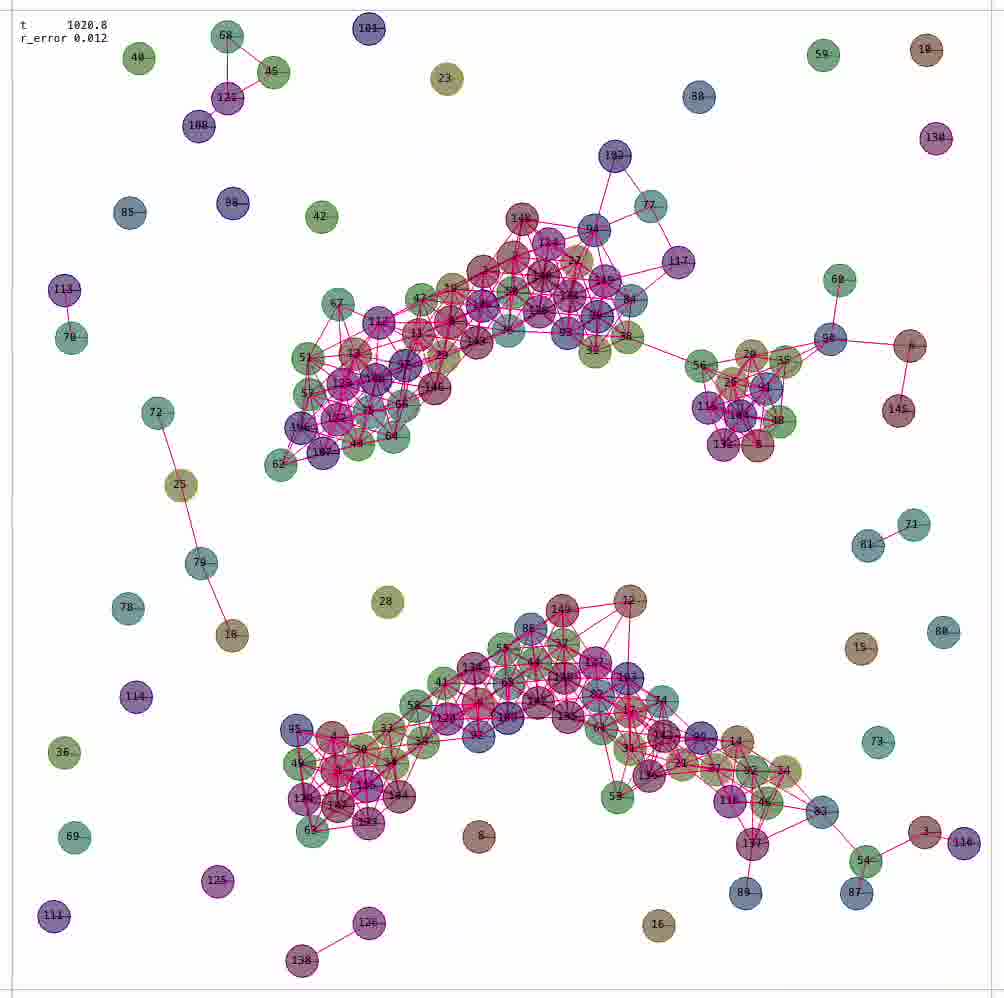}
	\includegraphics[width=0.16\linewidth]{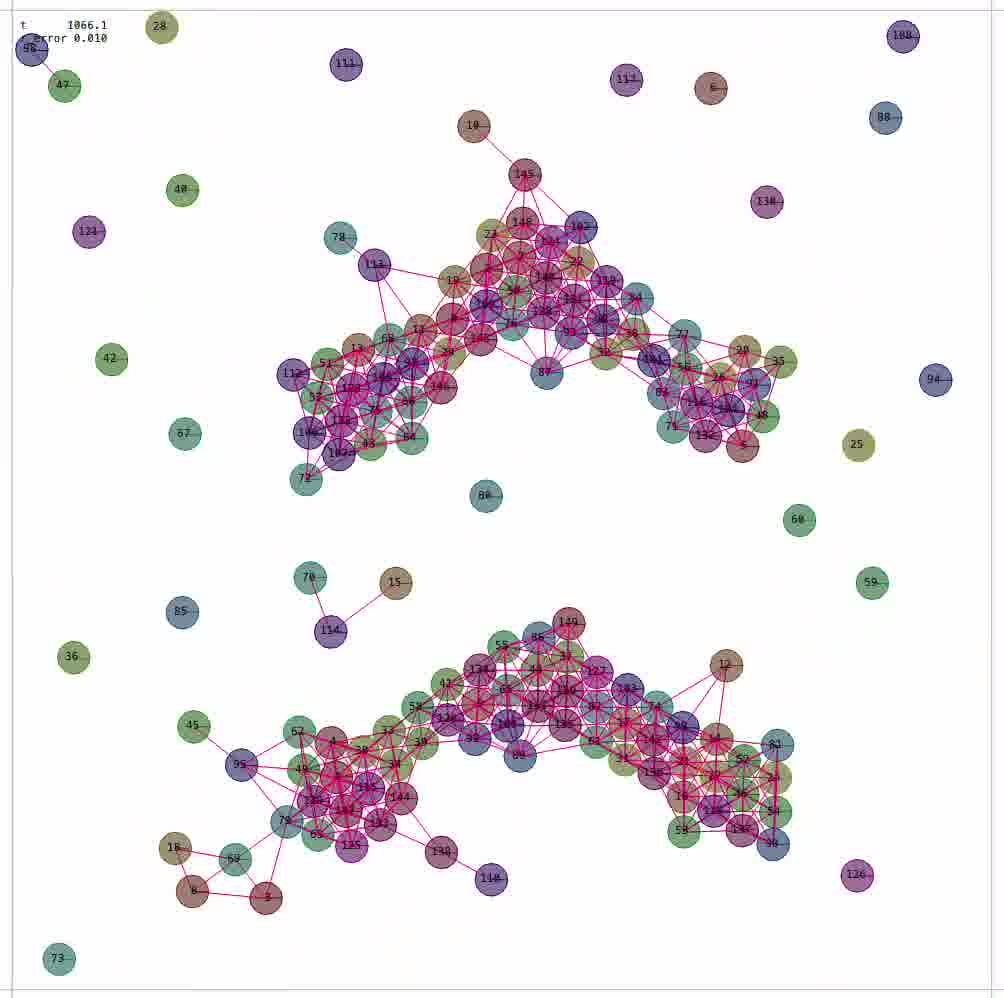}
	\includegraphics[width=0.16\linewidth]{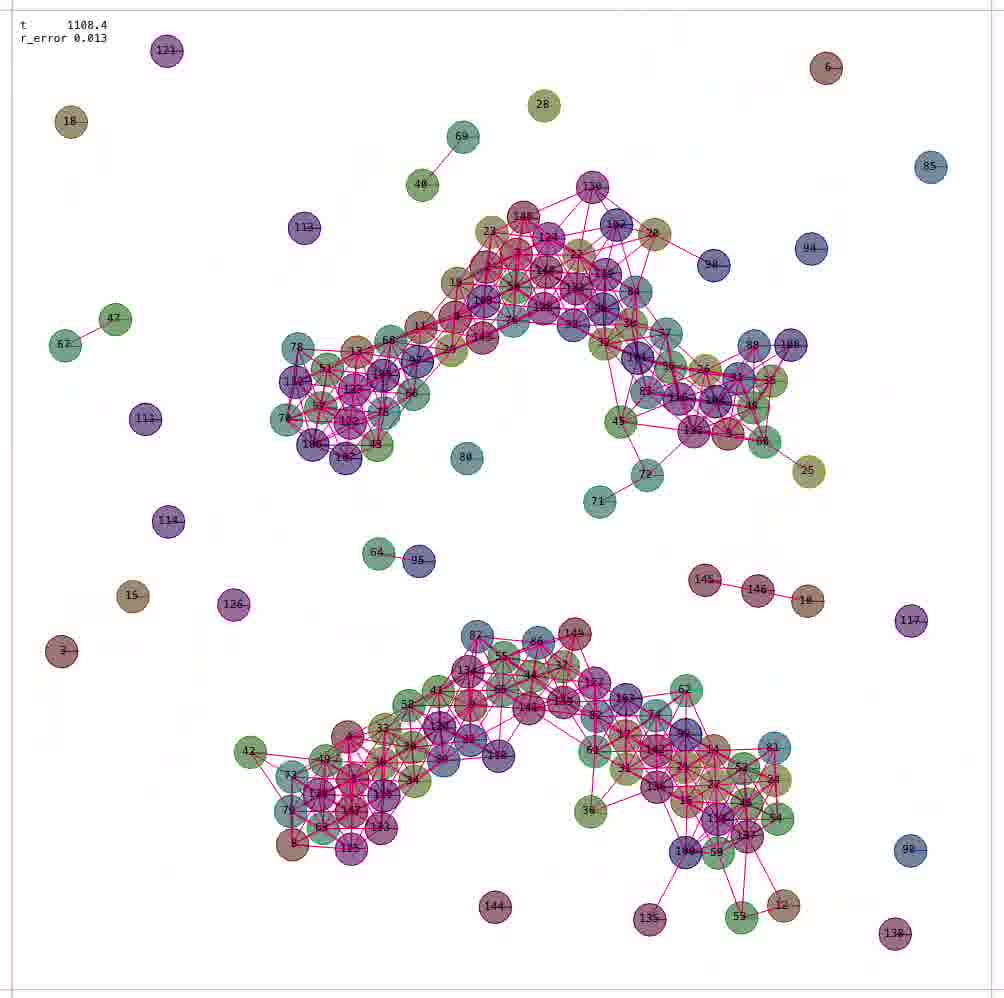}
	\includegraphics[width=0.16\linewidth]{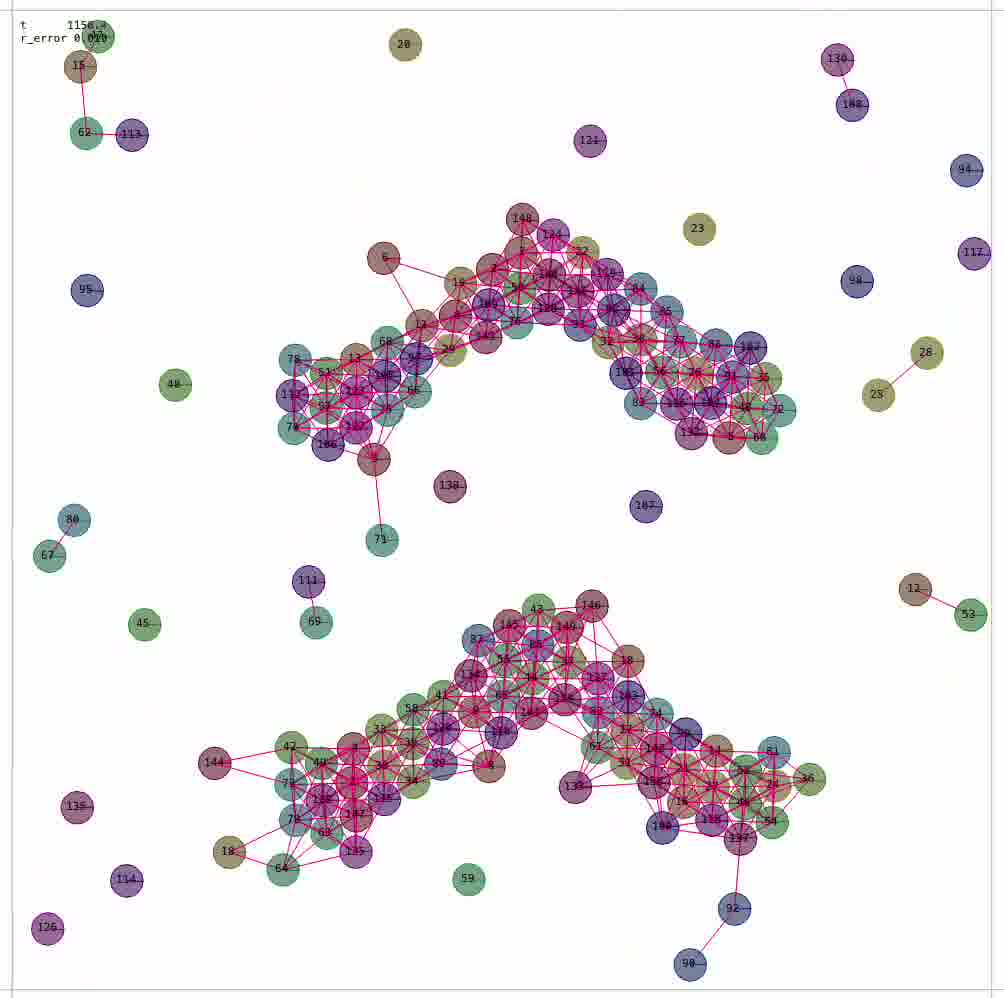}
	\includegraphics[width=0.16\linewidth]{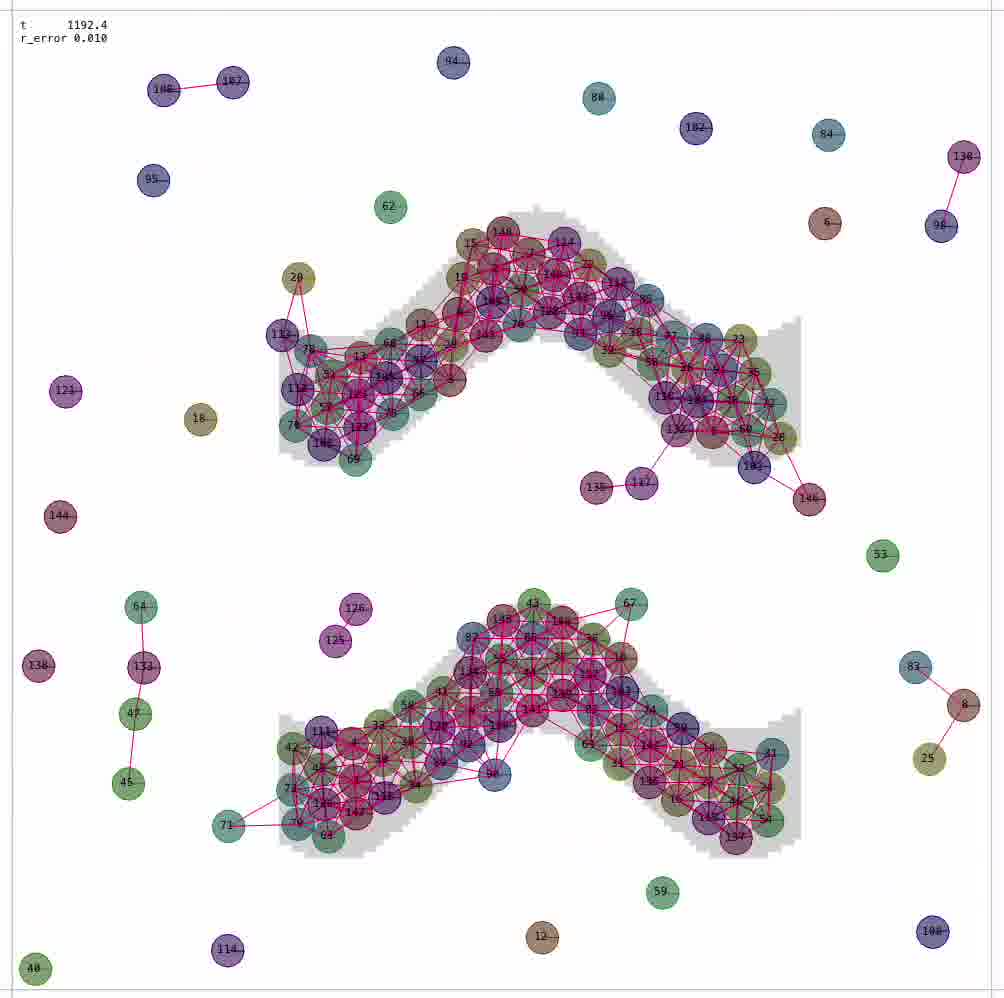}
\caption{Shape formation. 150 robots perform random walk DSA-RW when outside shape, and slow down and aggregate when inside shape, taking about 40 seconds to form each pattern.}
\label{fig:shape}
\end{figure}

We ran simulations using 150 robots, with an arena size of \SI{7.5}{m} per side. After reaching convergence, each robot followed Algorithm \ref{alg:shape}, with the shape function switching at regular intervals. Figure \ref{fig:shape} shows snapshots of the  process, with the desired shapes emerging within about 40 seconds in each case. The video at \url{https://youtu.be/ps5Wf-3UHr0} shows this process in full.

The algorithm is extremely simple, yet forms and cleanly transitions between shapes even without fine tuning of parameters.

\subsection{Intralogistics}
\begin{figure}
	\centering
	\includegraphics[trim={14 0 12 13},clip,width=0.49\linewidth]{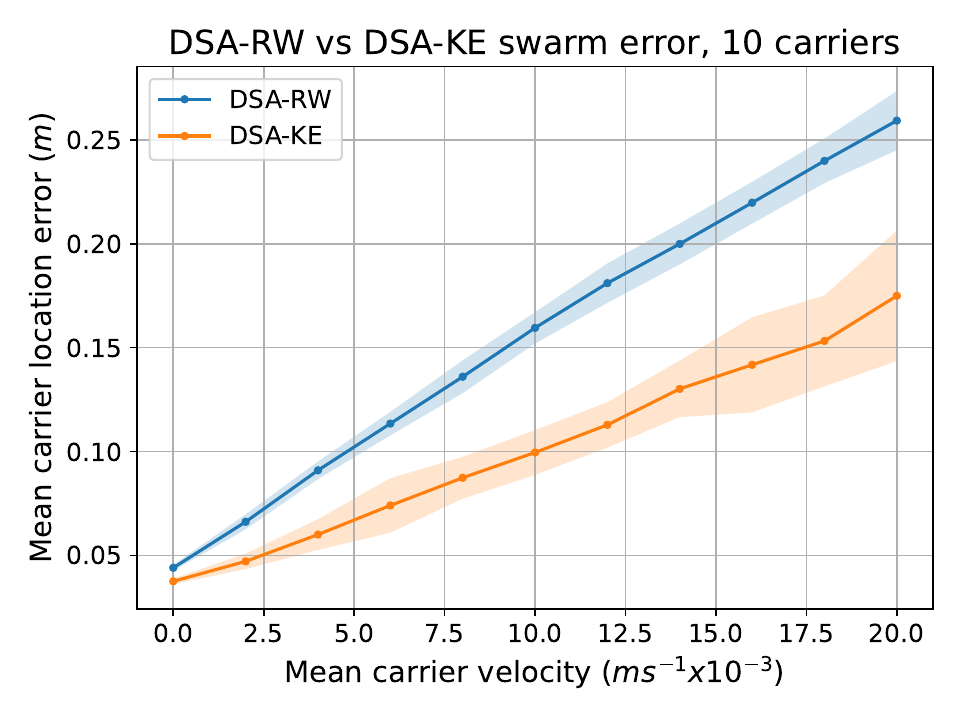}
	\includegraphics[trim={14 0 15 17},clip,width=0.495\linewidth]{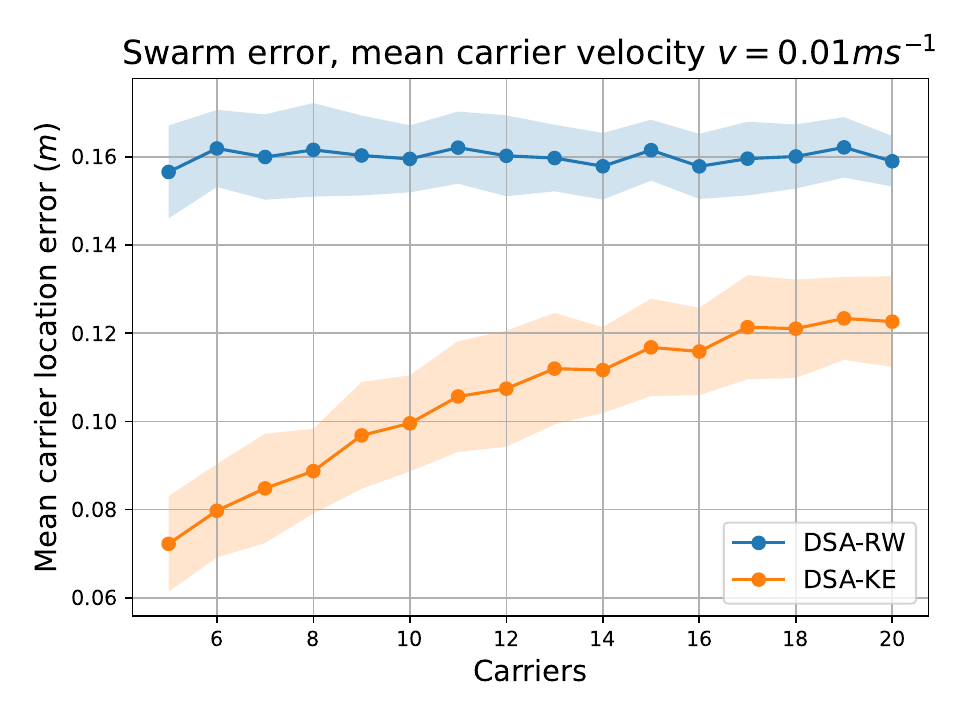}
\caption{Swarm carrier location error $s_{error}$ with the two robot behaviours while varying the carrier velocity (left) and the number of carriers in the arena (right). The number of robots is fixed at 10. Shaded areas indicate $\pm\sigma$ over 50 different random seeds. As the mean carrier velocity increases, so does the error, which is expected. With a fixed mean carrier velocity and sweeping the number of carriers, we see roughly constant error with the random walk of DSA-RW, but much lower error with DSA-KE as it actively seeks to enhance knowledge, particularly with low carrier numbers. In all cases, the DSA-KE behaviour results in much lower swarm perception error. }
\label{fig:error}
\end{figure}

Figure \ref{fig:error}  shows how the two movement behaviours DSA-RW and DSA-KE perform in acquiring knowledge about the dynamic carrier environment. At zero carrier velocities, the swarm quickly reaches low $s_{error}$ values, around \SI{0.04}{m}. Since the position sense has injected noise of $\sigma_{psense}=0.02m$, this is a reasonable lower bound. As the velocity of the carrier movement increases, the error goes up as unobserved carriers can move further from their last know positions. 

The effectiveness of DSA-KE in improving carrier position estimation in the swarm by actively moving to areas where knowledge is weaker, is clearly apparent. In all cases over different numbers of carriers and different carrier velocities, we see improvement, 38\% with 10 carriers and mean velocity \SI{0.01}{ms^{-1}}. When looking at performance vs the number of carriers, we see that DSA-RW is roughly constant, whereas DSA-KE performs extremely well with lower numbers of carriers. Clearly visible in the video available at \url{https://youtu.be/r53Z1O1pfxk} is a rather interesting emergent flocking behaviour; as one robot heads towards the carrier it knows least about, the backwards propagation of information from the leading robot on average causes the trailing robot to be in the best position to seek the next low-knowledge carrier, cohering the swarm. 
The good relative performance of DSA-KE falls off at higher numbers of carriers, so it may be that a better strategy for high carrier numbers would ensure a greater degree of dispersion.

It is important to note that both DSA-KE (Knowledge Enhancement) and DSA-SF (Shape Formation) are simple algorithms that serve to illustrate the possibilities that distributed spatial awareness can offer for swarm algorithms. The behaviour of the GBP message passing algorithm underlying the DSA shared reference frame is robust across parameters and the computational cost per robot of maintaining its local factor graph is low, of the order of a few hundred floating point operations per second, and the communications likewise is low cost, being a few hundred bytes per second. This is within reach of even cheap microcontrollers. 
But this low-cost outlay provides a rich, entirely decentralised, pseudo-global source of information not typically available to agents within swarms.

\section{Conclusion}\label{sec:conc}
We use exploratory movement, local observation, and distributed factor graph construction, combined with GBP evaluation, to give a global sensing ability to a swarm system without violating the core tenet of decentralisation, and with low per-robot computation and communication cost. 

Using two simple algorithms that make use of this new distributed spatial awareness ability, we show the possibilities. Applying some of the automatic design techniques commonly used for swarm controllers to systems using DSA is interesting; evolution often discovers non-obvious uses of new abilities. We consider GBP to be a powerful technique to infer global knowledge within a completely distributed paradigm, and wish to bring other items of state and knowledge within this overarching framework; it will be interesting to compare some more traditional swarm consensus algorithms with GBP, many best-of-n swarm algorithms rely on local message passing \cite{valentini2017best}, perhaps suggesting deeper similarities. 

We look forward to exploring new swarm algorithm possibilities that may combine DSA acquired knowledge with other swarm techniques. The current simulation system relies on graph construction happening in synchronisation at regular timesteps, but we don't see this as necessary and plan to remove this restriction, e.g. by estimating between states at time of observation. Finally, we are planning to implement the algorithm on the DOTS swarm in reality as part of our logistics demonstration task. 

\subsubsection{Acknowledgements} SJ and SH are funded by URKI grant 10038942 and EU grant 101070918.
%
%
%
\bibliographystyle{unsrtnat}


\end{document}